%% file: main.tex
\documentclass{article}

\usepackage{iclr2027_conference,times}
\usepackage{capt-of}
\usepackage{wrapfig}
\usepackage{amsmath,amssymb,mathtools}
\usepackage{booktabs}
\usepackage{multirow}
\usepackage{graphicx}
\usepackage{xcolor}
\usepackage{xspace}
\usepackage{microtype}
\usepackage{algorithm}
\usepackage[noend]{algpseudocode}
\usepackage{tikz}
\usetikzlibrary{arrows.meta,positioning,fit}
\usepackage{hyperref}
\usepackage{url}
\usepackage{longtable,array}

\definecolor{lessblue}{RGB}{35,83,138}
\definecolor{lesslight}{RGB}{235,243,250}
\definecolor{tbdred}{RGB}{165,35,38}
\newcommand{\method}{\textsc{LESS}\xspace}

\newcommand{\cma}{\textsc{CMA-ES}\xspace}

\title{LESS: Lightweight Evolutionary Supernet Search in Minutes}

\author{%
\parbox{0.96\textwidth}{%
\centering\normalfont
\textbf{Aviral Gandhi}$^{1}$ \quad
\textbf{Jinglue Xu}$^{2}$ \quad
\textbf{Jialong Li}$^{3}$ \quad
\textbf{Hitoshi Iba}$^{1}$\\[0.4em]
$^{1}$The University of Tokyo
\quad
$^{2}$Sakana AI
\quad
$^{3}$Waseda University\\[0.25em]
\texttt{aviral241202@gmail.com}
\quad
\texttt{jingluexu@sakana.ai}\\
\texttt{lijialong@fuji.waseda.jp}
\quad
\texttt{iba@iba.t.u-tokyo.ac.jp}
}}

\iclrfinalcopy

\begin{document}

\maketitle
\fancyhead{}

\begin{center}
\vspace{-1.25em}
\small
\url{https://github.com/AviralGandhi/LESS}
\end{center}

\begin{abstract}
Low-cost NAS must both explore high-performing architectures and identify them reliably, yet reducing evaluation cost often weakens the fidelity of candidate comparisons. Training-free methods reduce evaluation cost by replacing learned task feedback with proxy signals measured at initialization. We introduce LESS (Lightweight Evolutionary Supernet Search), a data-driven method that combines a brief fair hard-path warm-up with discrete search under a single CMA-ES distribution. Each proposal is evaluated as its decoded hard genotype after six candidate-conditioned supernet updates. On NAS-Bench-201, LESS achieves \(93.189\pm0.467\%\) CIFAR-10 test accuracy in 409.1 seconds, coming within \(0.04\) percentage points of FairNAS using approximately \(1/24\) of its source-reported search time. Matched controls show that calibration improves selected validation accuracy by \(0.577\) percentage points while changing best-visited accuracy by only \(0.054\) points, indicating that its primary effect is to reduce selection regret. The frozen configuration transfers without tuning to CIFAR-100 and ImageNet16-120 with \(69.615\pm1.139\%\) and \(43.720\pm1.697\%\) accuracy. Applied without tuning to the larger DARTS space, LESS achieves \(96.95\pm0.14\%\) on CIFAR-10 and \(82.43\pm0.80\%\) on CIFAR-100, with each search completing in approximately 43.5 minutes on a single GPU. Together, these results show that short, balanced, data-dependent updates enable competitive neural architecture search across datasets and search spaces within minutes.
\end{abstract}

\section{Introduction}
\label{sec:introduction}

Neural architecture search replaces manual network design with optimization over a structured family of networks. Early reinforcement-learning and evolutionary systems discovered competitive cells, but required hundreds to thousands of accelerator days because proposed networks were trained from scratch \citep{zoph2018learning,real2019regularized}. Weight-sharing NAS changed this cost profile: a single over-parameterized supernet contains many candidate subnetworks, allowing each candidate to reuse shared weights rather than train independently \citep{pham2018efficient,bender2018understanding,liu2019darts}. These methods reduce training cost, but they must solve two distinct problems: encountering strong architectures and reliably selecting them using shared-weight evaluations.

Two broad strategies improve low-cost candidate evaluations. Fair single-path methods train the supernet while balancing optimization across competing operations \citep{guo2020spos,chu2021fairnas}; training-free or ``zero-cost'' methods instead rank untrained networks using activation, gradient, Jacobian, neural-tangent-kernel, or composite signals \citep{mellor2021naswot,abdelfattah2021zerocost,chen2021tenas,li2023zico,lee2024aznas}. The former retain learned task feedback but require substantial supernet training and evaluation, 
whereas the latter search rapidly using handcrafted proxies whose rankings vary across search spaces and datasets \citep{krishnakumar2022nasbenchsuitezero}. Between these poles, low-fidelity estimators trade limited training for a more reliable signal, but typically require substantially more optimization or separate performance estimation for each candidate \citep{ru2021tse,white2021predictors}.

We investigate whether brief task-dependent training can improve architecture
selection while preserving a low search cost. We introduce LESS,
Lightweight Evolutionary Supernet Search, which adapts shared weights during
candidate evaluation. Building on balanced operation training
\citep{chu2021fairnas}, each proposed architecture participates in a small
group of complementary networks that balance operation exposure, while only
the proposed architecture is scored. Candidates start from the same weights
and use shared data, and only the winning candidate's updated state is
retained for subsequent search.

Across 24 paired NAS-Bench-201/CIFAR-10 searches
\citep{dong2020nasbench201}, calibration improves selected validation
accuracy by 0.577 percentage points while changing best-visited accuracy
by only 0.054 points. This indicates that the improvement primarily
reflects reduced selection regret---the gap between the best architecture
encountered and the architecture returned. LESS completes this search in
approximately seven minutes on one RTX~4090 and applies the same core
search settings across datasets and search spaces, including the DARTS
space \citep{liu2019darts}.

Our contributions are:
\begin{itemize}
    \item \textbf{Low-cost architecture evaluation.}
    We introduce candidate-conditioned calibration that combines brief,
    balanced training with direct evaluation of each proposed architecture.

    \item \textbf{An efficient search pipeline.}
    We integrate this evaluator into evolutionary search and demonstrate
    its application across datasets and search spaces without retuning
    the core search hyperparameters.

    \item \textbf{An analysis of architecture selection.}
    We distinguish the quality of architectures encountered from that
    of the architecture returned and show how lightweight calibration
    reduces selection regret.
\end{itemize}

\begin{figure*}[t]
    \centering
    \includegraphics[width=\textwidth]{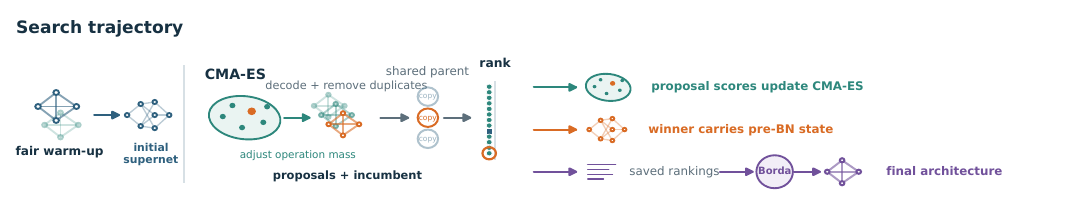}
    \caption{\textbf{LESS search trajectory.}
    A short fair hard-path warm-up initializes the supernet. At each search
    epoch, CMA-ES generates genotype-distinct proposals, which are evaluated
    alongside the incumbent from a shared parent state. Proposal losses update
    CMA-ES, while the epoch winner determines the persistent pre-BatchNorm
    supernet state. The final architecture is selected by aggregating rankings
    across the complete search trajectory using Borda scoring.}
    \label{fig:less-search}
\end{figure*}

\section{Related Work}
\label{sec:related}

\paragraph{Weight sharing and fair supernets.}
ENAS and one-shot NAS amortize candidate training through shared weights \citep{pham2018efficient,bender2018understanding}. SPOS instead trains and evaluates single paths, and Once-for-All and EagleEye recalibrate BatchNorm statistics before subnetwork evaluation \citep{guo2020spos,cai2020onceforall,li2020eagleeye}. FairNAS addresses unequal operation exposure through strictly balanced hard-path groups, while related methods partition the search space or concentrate training on promising candidates \citep{chu2021fairnas,zhao2021fewshot,hu2022generalizing,you2020greedynas}. Whereas Supernet Shifting accumulates updates into a single supernet that specializes toward frequently sampled regions \citep{zhang2024boosting}, \method applies fairness locally during candidate comparison: candidate proposals receive isolated balanced updates from a common parent state, and only the winner's state persists.

\paragraph{Low-cost evaluation and architecture selection.}
Training-free NAS ranks architectures using signals computed at or near initialization. NASWOT measures activation-pattern diversity, TE-NAS combines neural-tangent-kernel conditioning with linear-region counts, and subsequent methods use pruning saliency, gradient statistics, or ensembles of zero-cost signals \citep{mellor2021naswot,chen2021tenas,abdelfattah2021zerocost,li2023zico,lee2024aznas}. Low-fidelity estimators instead exchange limited training for stronger task feedback through early learning dynamics or learned performance predictors \citep{ru2021tse,white2021predictors}. DARTS-PT further shows that optimizing a supernet does not guarantee reliable architecture extraction \citep{wang2021rethinking}. \method addresses this selection problem using bounded task-dependent updates followed by evaluation of the exact hard genotype, and distinguishes the quality of the architectures visited from that of the architecture returned.

\paragraph{Evolutionary search.}
Evolutionary NAS directly optimizes discrete architectures but is expensive when every proposal is trained independently \citep{real2019regularized}. CMA-ES adapts a continuous distribution from ranked black-box evaluations, and CMANAS applies it to NAS using a pretrained one-shot model as a fixed fitness estimator \citep{hansen2001cmaes,sinha2024cmanas}. \method instead begins with a short warm-up and continues adapting temporary supernet states during candidate evaluation, combining CMA-ES exploration with candidate-conditioned fair calibration and persistent winner-state inheritance.

\begin{figure*}[t]
    \centering
    \includegraphics[width=\textwidth]{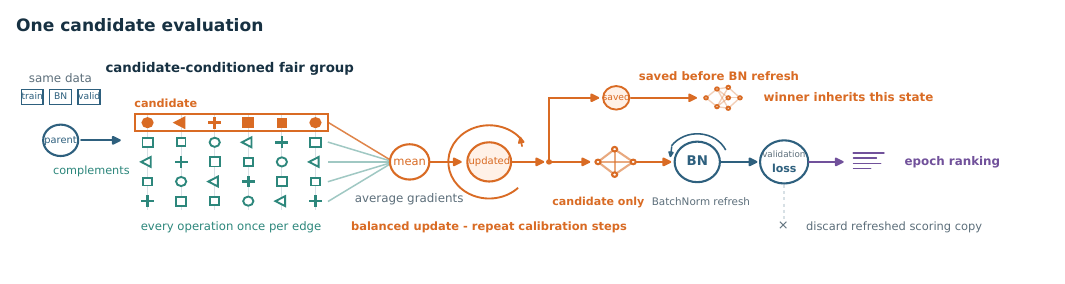}
    \caption{\textbf{Candidate-conditioned fair evaluation.}
    Each candidate anchors a group of complementary hard paths that jointly
    covers every eligible operation once per edge. Their gradients are averaged
    to produce one balanced supernet update, repeated across the calibration
    steps. Only the candidate architecture is scored after BatchNorm refresh.
    The refreshed scoring copy is discarded, while the saved pre-refresh state
    is inherited only if the candidate wins the epoch.}
    \label{fig:less-candidate-evaluation}
\end{figure*}
\section{Method}
\label{sec:method}

\method combines evolutionary search with candidate-conditioned
calibration from a shared parent state. Only the epoch winner's
calibrated state persists, and normalized Borda aggregation selects
the final architecture. Algorithm~\ref{alg:less} summarizes the procedure.

\begin{algorithm}[t]
\caption{\method}
\label{alg:less}
\small
\begin{algorithmic}[1]
\Require supernet $\Theta$, decoder $g$, population size $P$, calibration depth $K$, search epochs $T$
\State $\Theta_1\gets\Call{FairWarmup}{\Theta}$
\State initialize CMA-ES distribution and incumbent $a^{\mathrm{inc}}_1$
\For{$t=1,\ldots,T$}
    \State $\mathcal A_t\gets\Call{DistinctCMAProposals}{P}$ \quad
    \State$\mathcal Q_t\gets\{a^{\mathrm{inc}}_t\}\cup\mathcal A_t$
    \State materialize shared calibration, BN-refresh, and validation data
    \ForAll{$a\in\mathcal Q_t$}
        \State $g_a\gets\Call{Decode}{a}$;\quad
               $\Theta_{t,a}\gets\Call{Copy}{\Theta_t}$;\quad
               initialize a fresh optimizer
        \For{$k=1,\ldots,K$}
            \State $\mathcal G_{t,k}(g_a)\gets\Call{FairGroup}{g_a}$
            \State clear gradients
            \ForAll{$h\in\mathcal G_{t,k}(g_a)$}
                \State accumulate $\nabla_{\Theta}\mathcal L_{\mathrm{tr}}
                (h;\Theta_{t,a},\mathcal B^{\mathrm{cal}}_{t,k})
                /|\mathcal G_{t,k}|$
            \EndFor
            \State perform one SGD update of $\Theta_{t,a}$

        \EndFor
        \State $\bar{\Theta}_{t,a}\gets\Call{Copy}{\Theta_{t,a}}$
               \Comment{state eligible for inheritance}
        \State $\Theta^{\mathrm{score}}_{t,a}
    \gets \Call{Copy}{\bar{\Theta}_{t,a}}$
\State $\Theta^{\mathrm{score}}_{t,a}
    \gets \Call{BNRefresh}{\Theta^{\mathrm{score}}_{t,a},
    g_a,\mathcal B^{\mathrm{bn}}_t}$
    \Comment{refresh only the scoring copy}
        \State $\ell_t(a)\gets
               \Call{HardPathLoss}{g_a,\Theta^{\mathrm{score}}_{t,a},
               \mathcal B^{\mathrm{val}}_t}$
    \EndFor
    \State store the within-epoch ranking of $\mathcal Q_t$; update CMA-ES using
           $\{(a,\ell_t(a)):a\in\mathcal A_t\}$
    \State $a_t^\star\gets\arg\min_{a\in\mathcal Q_t}\ell_t(a)$;\quad
           $(\Theta_{t+1},a^{\mathrm{inc}}_{t+1})
           \gets(\bar{\Theta}_{t,a_t^\star},a_t^\star)$
\EndFor
\State \Return $\Call{BordaSelect}{\text{stored rankings}}$
\end{algorithmic}
\end{algorithm}

\subsection{Architecture representation and decoding}

Let $E$ denote the number of searchable edges and $\mathcal O_e$ the operation set on edge $e$, with $O=|\mathcal O_e|$. CMA-ES operates on edge-operation logits $a\in\mathbb R^{E\times O}$, which define probabilities $p_e=\operatorname{softmax}(a_e)$. Unlike learned continuous encodings \citep{luo2018nao}, these variables directly parameterize the discrete operation choices.

Short shared-weight evaluations can favor inexpensive parameter-free operations, particularly \texttt{none} and \texttt{skip}, consistent with collapse modes observed in differentiable NAS \citep{chu2020fairdarts,zela2020understanding}. Our development runs similarly showed that uncorrected proxy rankings concentrated on architectures containing these operations. We therefore apply an operation-level mass redistribution before decoding. For the available operation set, let $\mathcal D=\mathcal O_e\cap\{\texttt{none},\texttt{skip}\}$ contain the donor operations and $\mathcal R=\mathcal O_e\setminus\mathcal D$ the remaining operations. We define
\begin{equation}
\tilde p_{e,o}=
\begin{cases}
(1-r_o)p_{e,o}, & o\in\mathcal D,\\[1mm]
p_{e,o}+\displaystyle\frac{1}{|\mathcal R|}
\sum_{d\in\mathcal D}r_dp_{e,d}, & o\in\mathcal R,
\end{cases}
\label{eq:redistribution}
\end{equation}
using the fixed rates $r_{\texttt{none}}=0.55$ and $r_{\texttt{skip}}=0.50$. The removed mass is distributed uniformly over the remaining operations, preserving total probability. These rates were set once using NAS-Bench-201/CIFAR-10 development experiments and held fixed across all subsequent datasets and search spaces. The rule assumes that \texttt{none} and \texttt{skip} retain their standard null and identity semantics; adapting this prior to macro-level and transformer search spaces remains future work.

The deterministic decoder $g(\tilde p)$ then applies the canonical topology constraints of the relevant search space to produce a valid discrete genotype. Because multiple continuous samples can decode to the same genotype, proposals are deduplicated after decoding. At each epoch, CMA-ES draws until it obtains $P=12$ distinct, non-incumbent genotypes. If the bounded proposal budget is exhausted, a deterministic fallback fills the remaining slots while preserving the population size. Draws excluded from the final population are neither evaluated nor used to update CMA-ES.

\subsection{Candidate-conditioned fair calibration}
\label{sec:calibration}

Before search, we pretrain the supernet for 25 epochs using balanced groups of discrete hard paths, without optimizing architecture variables. This produces the initial parent state $\Theta_1$; the warm-up is included in the reported end-to-end search cost.

At epoch $t$, candidates start from separate copies of the same parent
state $\Theta_t$ and share calibration, BN-refresh, and validation data.

For NAS-Bench-201, consider a decoded candidate $g$. At calibration step $k$, we construct
$\mathcal G_{t,k}(g)=\{h_{k,1},\ldots,h_{k,O}\}$ with $h_{k,1}=g$, such that
$\{h_{k,j}(e):j=1,\ldots,O\}=\mathcal O_e$ for every edge $e$.
Thus, the candidate is one complete path in the group, while the remaining paths combine the operations not selected by the candidate. Every eligible operation is executed exactly once on every edge. This adapts the strict-fairness principle of FairNAS \citep{chu2021fairnas} to candidate-conditioned evaluation.

For example, if an NB201 candidate uses a $3\times3$ convolution on an
edge, the four complementary paths use the remaining operations on that
edge: \texttt{none}, skip connection, $1\times1$ convolution, and average
pooling. Repeating this construction across all edges produces five
complete paths whose gradients are averaged for one update; only the
original candidate is subsequently scored. DARTS \citep{liu2019darts}
instead uses approximate balance under its topology constraints, as
detailed in Appendix~\ref{app:darts-adapter}.

All paths in the group are evaluated sequentially on the same calibration minibatch. Their gradients are accumulated before a single optimizer step:
\begin{equation}
\mathcal L^{\mathrm{fair}}_{t,k}
=
\frac{1}{O}
\sum_{h\in\mathcal G_{t,k}(g)}
\mathcal L_{\mathrm{tr}}
\left(h;\Theta^{k-1}_{t,g},
\mathcal B^{\mathrm{cal}}_{t,k}\right),
\label{eq:fairloss}
\end{equation}
\begin{equation}
\Theta^k_{t,g}
=
\operatorname{SGD}
\left(
\Theta^{k-1}_{t,g},
\nabla_{\Theta}\mathcal L^{\mathrm{fair}}_{t,k}
\right).
\label{eq:fairupdate}
\end{equation}
The candidate therefore contributes a gradient in the context of its exact topology, while the complementary paths prevent the temporary supernet from being optimized exclusively toward that candidate.

The frozen protocol uses $K=6$. NAS-Bench-201 has five operations per edge, so evaluating one candidate requires six optimizer steps and $6\times5=30$ hard-path forward/backward passes. The architecture itself remains fixed throughout calibration; only the temporary supernet state changes.

\subsection{Hard-path scoring, inheritance, and final selection}

After calibration, we retain the pre-refresh state
$\bar{\Theta}_{t,g}=\Theta^K_{t,g}$. A separate copy resets and re-estimates its
BatchNorm statistics using only the focal hard genotype:
$\Theta^{\mathrm{score}}_{t,g}
=\operatorname{BNRefresh}(\bar{\Theta}_{t,g},g,\mathcal B^{\mathrm{bn}}_t)$.
The candidate is then scored by mean validation cross-entropy,
\begin{equation}
\ell_t(g)=
\frac{1}{|\mathcal B^{\mathrm{val}}_t|}
\sum_{(x,y)\in\mathcal B^{\mathrm{val}}_t}
\operatorname{CE}\!\left(
f_g(x;\Theta^{\mathrm{score}}_{t,g}),y
\right).
\label{eq:score}
\end{equation}
For a candidate vector $a$, we use $\ell_t(a)$ as shorthand for $\ell_t(g(a))$, the validation loss of its decoded architecture.
The complementary paths affect the balanced calibration update but never enter
this score. The refreshed BatchNorm state is discarded after evaluation.

CMA-ES uses the losses of the $P$ sampled proposals to update its mean,
covariance, and scale. Separately, the best member of the complete population,
including the incumbent, determines the continuing search state. If
$\mathcal Q_t$ denotes the incumbent and proposal vectors, its lowest-loss member $a_t^\star=\arg\min_{a\in\mathcal Q_t}\ell_t(a)$ becomes the next incumbent, $a^{\mathrm{inc}}_{t+1}=a_t^\star$, while its pre-refresh calibrated state becomes the next parent, $\Theta_{t+1}=\bar{\Theta}_{t,g(a_t^\star)}$.

The epoch winner controls the next search state, but it is not necessarily the
architecture ultimately returned. Let $\mathcal U_t$ denote the unique
genotypes evaluated at epoch $t$. Duplicate occurrences are collapsed by their
mean loss, after which $r_t(g)$ denotes the within-epoch rank of genotype $g$,
with rank one being best. Let $m_t=|\mathcal U_t|$. The final architecture maximizes the cumulative
normalized Borda score:
\begin{equation}
g_{\mathrm{out}}=\arg\max_g\sum_{t:\,g\in\mathcal U_t}b_t(g),
\qquad
b_t(g)=
\begin{cases}
1, & m_t=1,\\
\dfrac{m_t-r_t(g)}{m_t-1}, & m_t>1.
\end{cases}
\label{eq:borda}
\end{equation}
Tie-breaking is specified in Appendix~\ref{app:borda-exact}.

\section{Results}
\label{sec:results}

\subsection{Evaluation setup}
\label{sec:experiments}

We evaluate \method on NAS-Bench-201, which contains 15,625 cells and provides standardized results for CIFAR-10, CIFAR-100, and ImageNet16-120 \citep{dong2020nasbench201}. Training data support warm-up, calibration, and BatchNorm refresh; disjoint validation data score candidates. The frozen protocol uses 25 warm-up epochs, six calibration updates, 12 proposals plus an incumbent, and 60 search epochs, for 780 candidate evaluations per run. CIFAR-10 includes a 16-seed paired cohort and an eight-seed prospective extension; each $K=0/K=6$ pair shares its warm-up checkpoint. We also run the unchanged $K=6$ search on eight seeds each for CIFAR-100 and ImageNet16-120. Held-out search validation data determine the returned architecture;
NAS-Bench-201 validation and test accuracies are joined only after search. All search times are measured on one RTX~4090. Full configurations appear in Appendix~\ref{app:config}.

We also evaluate DARTS-space transfer \citep{liu2019darts}, changing
the network and search-space adapter while retaining the core search
hyperparameters and selection rule. For each CIFAR dataset, we report
the mean and standard deviation of best test accuracy over 600
retraining epochs across three independently searched architectures
(Appendix~\ref{app:darts-results}).

\subsection{NAS-Bench-201 accuracy and search cost}

Across 24 paired CIFAR-10 search seeds, \method obtains
$89.977\pm0.597\%$ validation accuracy and $93.189\pm0.467\%$ test accuracy.
The matched $K=0$ control, which retains BatchNorm refresh but performs no calibration updates, obtains $89.401\pm0.584\%$ and $92.693\pm0.431\%$, respectively. The paired validation gain is $0.529$ points in the initial 16-seed cohort and $0.671$ points in the prospective eight-seed extension, yielding $0.577$ points across all 24 pairs; 19 pairs favor $K=6$.

Table~\ref{tab:nb201-comparison} compares \method with source-reported
NAS-Bench-201 results. Its test accuracy is within $0.041$ percentage points of FairNAS,
with source-reported search times of 409 and 9,845 seconds, respectively. Compared with
NASWOT using 1,000 samples, \method improves test accuracy by $0.229$ points
and reduces the reported standard deviation from $0.81$ to $0.47$. TE-NAS
obtains $0.711$ points higher accuracy but requires $3.8\times$ more reported
wall-clock time. These source-reported times use different hardware and implementations and therefore do not establish hardware-normalized speedups. To make the computational workload explicit, LESS performs 4,680 calibration
updates per default search, comprising 23,400 hard-path
forward/backward passes on NB201 and 32,760 on DARTS. These calibration
counts exclude warm-up, BatchNorm refresh, validation, and final
retraining; Appendices~C.5 and~P provide the complete workload and
timing breakdowns.

\begin{table*}[t]
\caption{\textbf{NAS-Bench-201 accuracy--cost comparison.}
All accuracy values are test percentages. Times are source-reported and use
different hardware and implementations. For LESS, the reported times on
CIFAR-10, CIFAR-100, and ImageNet16-120 are 409.1, 408.9, and 453.1 seconds,
respectively.}
\label{tab:nb201-comparison}
\centering
\scriptsize
\setlength{\tabcolsep}{2pt}
\renewcommand{\arraystretch}{0.96}

\resizebox{0.98\textwidth}{!}{%
\begin{tabular}{@{}llrrrrr@{}}
\toprule
Method & Search class & CIFAR-10 & CIFAR-100 & ImageNet16-120
& Time (s) & Runs \\
\midrule
REA \citep{real2019regularized,mellor2021naswot}
    & evolution
    & $93.92\pm0.30$ & $71.84\pm0.99$ & $45.54\pm1.03$
    & -- & 500 \\
Random search \citep{mellor2021naswot}
    & training
    & $93.70\pm0.36$ & $71.04\pm1.07$ & $44.57\pm1.25$
    & -- & 500 \\
BOHB \citep{falkner2018bohb,mellor2021naswot}
    & multi-fidelity
    & $93.61\pm0.52$ & $70.85\pm1.28$ & $44.42\pm1.49$
    & -- & 500 \\
RSPS \citep{li2019random,mellor2021naswot}
    & weight sharing
    & $87.66\pm1.69$ & $58.33\pm4.34$ & $31.14\pm3.88$
    & 7,587 & 3 \\

ENAS \citep{pham2018efficient,mellor2021naswot}
    & weight sharing
    & $54.30$ & $15.61$ & $16.32$
    & 13,315 & 3 \\
SETN \citep{mellor2021naswot}
    & weight sharing
    & $86.19\pm4.63$ & $56.87\pm7.77$ & $31.90\pm4.07$
    & 31,010 & 3 \\
GDAS \citep{dong2019gdas,mellor2021naswot}
    & weight sharing
    & $93.51\pm0.13$ & $70.61\pm0.26$ & $41.84\pm0.90$
    & 28,926 & 3 \\
DARTS-V1 \citep{liu2019darts,mellor2021naswot}
    & differentiable
    & $54.30$ & $15.61$ & $16.32$
    & 10,890 & 3 \\
DARTS-V2 \citep{liu2019darts,mellor2021naswot}
    & differentiable
    & $54.30$ & $15.61$ & $16.32$
    & 29,902 & 3 \\
NASWOT, $N=10$ \citep{mellor2021naswot}
    & training-free
    & $92.44\pm1.13$ & $68.62\pm2.04$ & $41.31\pm4.11$
    & 3.05 & 500 \\
NASWOT, $N=100$ \citep{mellor2021naswot}
    & training-free
    & $92.81\pm0.99$ & $69.48\pm1.70$ & $43.10\pm3.16$
    & 30.01 & 500 \\
NASWOT, $N=1000$ \citep{mellor2021naswot}
    & training-free
    & $92.96\pm0.81$ & $69.98\pm1.22$ & $44.44\pm2.10$
    & 306.19 & 500 \\
TE-NAS \citep{chen2021tenas}
    & training-free
    & $93.90\pm0.47$ & $71.24\pm0.56$ & $42.38\pm0.46$
    & 1,558 & 4 \\
FairNAS \citep{chu2021fairnas}
    & fair supernet
    & $93.23\pm0.18$ & $71.00\pm1.46$ & $42.19\pm0.31$
    & 9,845 & 3 \\
\midrule
\method ($K=6$)
    & calibrated supernet
    & $93.189\pm0.467$ & $69.615\pm1.139$ & $43.720\pm1.697$
    & 409.1 & $24/8/8$ \\
\bottomrule
\end{tabular}%
}
\end{table*}

Across eight matched seeds, replacing candidate-conditioned calibration with candidate-independent balanced groups at the same update budget and with winner-state inheritance retained yielded $89.031\pm2.131\%$ selected validation accuracy, compared with $90.041\pm0.482\%$ for LESS. Although statistically inconclusive, the improvement was concentrated in two large selection failures of the independent control, suggesting that candidate conditioning may help avoid occasional severe mis-selection.

The archive decomposition explains where calibration helps. Best-visited validation accuracy changes only from $90.676\%$ at $K=0$ to $90.730\%$ at $K=6$, whereas selection regret falls from $1.275$ to $0.753$ points. The calibrated trajectories evaluate fewer distinct genotypes on average ($124.9$ versus $147.0$), but contain more candidate occurrences above the global 95th percentile ($16.79$ versus $11.04$) and 99th percentile
($4.25$ versus $0.63$). Archive-wide Spearman correlation between cumulative Borda scores and
benchmark validation accuracy increases from $0.463$ to $0.495$. Calibration therefore reduces selection regret and produces more upper-tail candidate occurrences despite visiting fewer distinct genotypes.
The gain is not explained solely by additional wall-clock time. On eight matched seeds, extending $K=0$ search to 180 epochs requires $529.0$ seconds and returns $89.397\%$ validation accuracy. The corresponding $K=6$ method requires $410.2$ seconds and returns $90.041\%$, with a shorter measured runtime.
\subsection{Component ablations}
\label{sec:ablations}

\begin{figure}[t]
    \centering
    \includegraphics[width=0.49\linewidth]{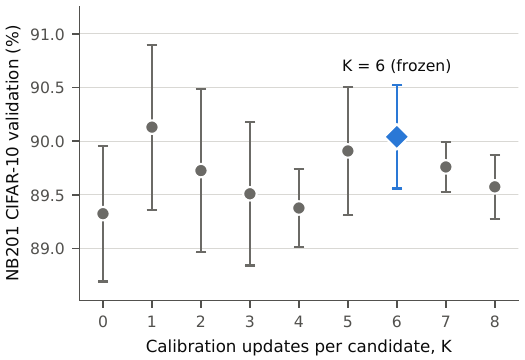   }
    \hfill
    \includegraphics[width=0.49\linewidth]{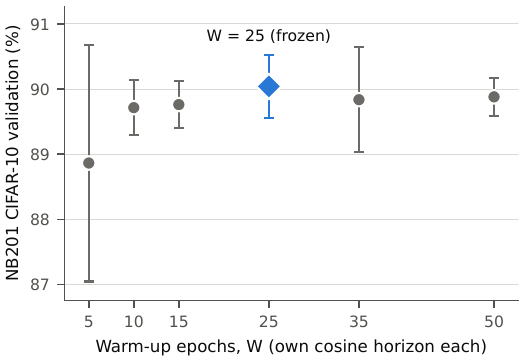}
    \caption{\textbf{Calibration and warm-up sensitivity.}
    Mean selected validation accuracy over eight matched seeds;
    error bars show one standard deviation. The frozen $K=6$ and
    $W=25$ settings are highlighted.}
    \label{fig:kw-ablation}
\end{figure}

Figure~\ref{fig:kw-ablation} shows sensitivity to calibration
depth and warm-up across eight matched ablation seeds. With $W=25$, introducing one calibration update raises mean
selected validation accuracy from $89.325\%$ at $K=0$ to $90.131\%$ at $K=1$.
Performance then varies non-monotonically: the frozen $K=6$ setting reaches
$90.041\%$, only $0.090$ points below $K=1$. Although $K=1$ is faster
(243 versus 410 seconds), $K=6$ has lower across-seed variation
($0.482$ versus $0.769$ points) and lower selection regret
($0.623$ versus $0.830$ points). The curve supports the value of calibration,
but neither increasing returns with $K$ nor a unique optimum at $K=6$.

With $K=6$ fixed, $W=25$ has the highest mean selected validation accuracy
($90.041\%$). A five-epoch warm-up is less reliable
($88.865\pm1.817\%$), while extending warm-up to 35 or 50 epochs adds cost
without improving the mean. Borda--truth Spearman correlation rises from
$0.187$ at $W=5$ to $0.525$ at $W=25$, with no further gain at $W=50$.
Each warm-up length uses its own cosine schedule with $T_{\max}=W$; these
points compare complete training recipes, not prefixes of one trajectory.
The predefined $W25/K6$ configuration remains fixed for the main results.

Replacing CMA-ES with uniform discrete proposals expands the archive from
$132.9$ to $704.5$ unique genotypes and raises best-visited validation accuracy
from $90.663\%$ to $91.389\%$, yet lowers selected accuracy from $90.041\%$ to
$89.010\%$ and increases regret from $0.623$ to $2.379$ points. Matching the
uniform archive to $132.9$ genotypes offline does not recover the gap:
selected accuracy remains $88.92\%$, with CMA-ES winning on all eight seeds.
Under the same low-fidelity evaluation procedure, CMA-ES returns better
architectures than uniform proposals despite visiting fewer distinct
genotypes.

Removing mass redistribution similarly increases the best accuracy visited
but reduces selected validation accuracy from $90.041\%$ to $89.493\%$.
This observed improvement supports the use of mass redistribution as an
operation-level prior in the evaluated setting.

Replaying alternative selectors on the 16-seed $K=6$ archive produces
$89.848\%$ validation accuracy for Borda, $89.881\%$ for the final-epoch input incumbent,
and $89.955\%$ for the last-epoch winner. The differences are small and not
statistically significant in this cohort, indicating that the three selectors
perform comparably on average. We retain Borda as a fixed trajectory-wide
rule that aggregates rankings across the complete search rather than relying
on a single final comparison.
\subsection{Dataset transfer}
\label{sec:cifar100}

\begin{figure}[t]
    \centering
    \includegraphics[width=0.49\linewidth]{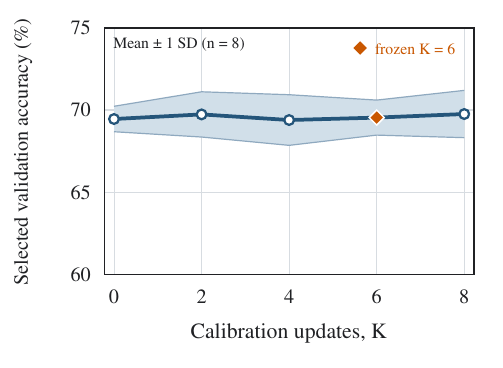}
    \hfill
    \includegraphics[width=0.49\linewidth]{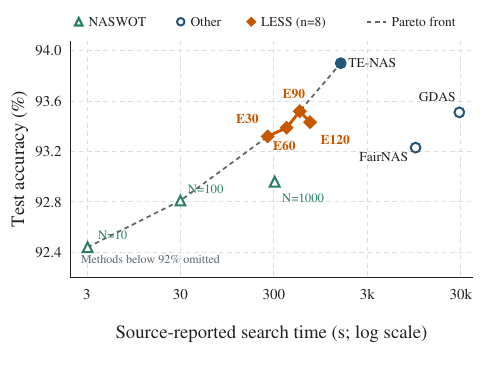}
    \caption{\textbf{CIFAR-100 calibration depth and accuracy--compute trade-off.}
    Left: mean CIFAR-100 validation accuracy over eight seeds, with
    one-standard-deviation error bars. Right: LESS E30--E120 prefixes and
    source-reported efficient NAS baselines; times use different hardware
    and implementations.}
    \label{fig:c100-k}
    \label{fig:pareto}
\end{figure}

Without retuning the CIFAR-10 search hyperparameters, \method obtains
$69.538\pm1.064\%$ validation accuracy and $69.615\pm1.139\%$ test accuracy
on CIFAR-100 in 408.9 seconds. Figure~\ref{fig:c100-k} shows the corresponding
calibration-depth study. Although the highest observed mean is $69.755\%$ at
$K=8$, the means across $K\in\{0,2,4,6,8\}$ range only from $69.390\%$ to
$69.755\%$, and every difference is smaller than one within-setting standard
deviation at $n=8$. The sweep therefore provides no clear evidence for
dataset-specific selection of a different calibration depth, and we retain the
predefined $K=6$ configuration as the transfer result.

Table~\ref{tab:nb201-comparison} provides published baselines;
NASWOT uses CIFAR-10-scored architectures, so these cross-protocol
comparisons are contextual rather than controlled.

We additionally apply the same frozen search hyperparameters to
ImageNet16-120. Across eight search seeds, \method obtains
$43.180\pm1.759\%$ validation accuracy and $43.720\pm1.697\%$ test accuracy
in 453.1 seconds. The selected architectures have a mean full-space percentile of 91.1.
Every run visits an architecture above the 99th percentile, but the
mean selection regret is 2.662 percentage points
(Appendix~\ref{app:imagenet}). These results indicate that selecting
among the high-performing architectures encountered remains the
principal limitation.

\subsection{Accuracy--compute scaling}
\label{sec:scaling}

Figure~\ref{fig:pareto} presents the accuracy--compute trajectory obtained by
replaying Borda selection after 30, 60, 90, and 120 search epochs. Mean
validation accuracy increases from $89.976\%$ at 257 seconds to $90.203\%$ at
567 seconds, with selection regret falling from $0.679$ to $0.525$ points. At 120 epochs, mean selected validation accuracy is $90.183\%$, slightly below the $90.203\%$ reached at 90 epochs despite the additional compute.

Late-search concentration explains this saturation. The number of new
genotypes discovered per 30-epoch block decreases from $95.5$ to $37.4$,
$12.8$, and $7.9$, while the number of raw CMA-ES draws needed to construct
distinct populations rises sharply. Additional compute still expands the
archive, but increasingly little of it changes the final selection.

\subsection{DARTS-space transfer}
\label{sec:darts}

Table~\ref{tab:darts} evaluates search-space transfer. Using the search
hyperparameters frozen on NAS-Bench-201, \method obtains
$96.95\pm0.14\%$ CIFAR-10 test accuracy across three independently searched
architectures. Each search requires 0.030 GPU-days on one RTX~4090. The result
is close to the accuracy of established DARTS-space methods while using a
smaller search budget and no DARTS-specific hyperparameter tuning.

\begin{table}[t]
\caption{\textbf{DARTS-space comparison.}
Test errors are percentages. Search costs follow the respective sources and
are not hardware-normalized. All comparison methods use configurations
developed for the DARTS space, whereas LESS transfers its frozen
NAS-Bench-201 search hyperparameters without retuning. LESS reports the mean
and standard deviation across three independently searched architectures per
dataset. A dash denotes an unavailable canonical CIFAR-100 result.}
\label{tab:darts}
\centering
\scriptsize
\setlength{\tabcolsep}{2.5pt}
\begin{tabular}{@{}lcccc@{}}
\toprule
Method & C10 Error (\%) & C100 Error (\%) & Params (M) & GPU-days \\
\midrule
DARTS-2 \citep{liu2019darts}
    & $2.76\pm0.09$  & $17.54$          & 3.3 & 4.00 \\
SNAS \citep{xie2019snas}
    & $2.85\pm0.02$  & --               & 2.8 & 1.50 \\
GDAS (FRC) \citep{dong2019gdas}
    & $2.82$         & $18.13$          & 2.5 & 0.17 \\
P-DARTS \citep{chen2019pdarts}
    & $2.50$         & $17.20$          & 3.4 & 0.30 \\
PC-DARTS \citep{xu2020pcdarts}
    & $2.57\pm0.07$  & --               & 3.6 & 0.10 \\
TE-NAS \citep{chen2021tenas}
    & $2.63\pm0.064$ & --               & 3.8 & 0.05 \\
RSWS \citep{li2019random}
    & $2.85\pm0.08$  & --               & 4.3 & 2.70 \\
\midrule
\method
    & $3.05\pm0.14$  & $17.57 \pm 0.80$ & 3.3 & 0.030 \\
\bottomrule
\end{tabular}

\vspace{1mm}
\parbox{\linewidth}{\scriptsize
For LESS, the parameter count is averaged over the three CIFAR-10
architectures and excludes the auxiliary classifier, following the
DARTS convention \citep{liu2019darts}.
The reported DARTS and P-DARTS CIFAR-100 values evaluate architectures
searched on CIFAR-10; LESS searches independently on each dataset.
RSWS reports 2.7 GPU-days for weight-sharing search and a further
6 GPU-days for architecture selection.}
\end{table}
On CIFAR-100, the same frozen search configuration obtains
$82.43\pm0.80\%$ test accuracy across three independently searched
architectures. Each search takes approximately 43.5 minutes
($0.030$ GPU-days) on one RTX 4090. These experiments evaluate
direct transfer rather than performance after search-space-specific tuning.

\section{Discussion}
\label{sec:discussion}

Calibration primarily improves selection: relative to $K=0$, selected
validation accuracy increases by $0.577$ percentage points, while
best-visited accuracy changes by only $0.054$ points. Alternative
selectors perform similarly on the recorded trajectories, so this
gain cannot be attributed to Borda alone.

Uniform proposals discover stronger architectures but return weaker
ones than CMA-ES under the same evaluator. Conversely, CMA-ES discovers
few new genotypes late in search, suggesting a need to renew exploration
while preserving reliable candidate comparisons.

Calibration also changes the structure of the selected NB201 cells.
Across the 24 paired CIFAR-10 searches, the mean number of convolutional
edges increases from 4.00 to 5.13 out of six, while skip connections
decrease from 1.17 to 0.71 and null edges from 0.38 to 0.13
(Appendix~E.1). These descriptive differences characterize the returned
architectures without establishing which structural changes cause the
accuracy improvement. Complete NB201 and DARTS genotypes are provided
in Appendices~M and~O.2.

The transfer experiments show that the frozen search configuration remains usable across three NAS-Bench-201 datasets and the larger DARTS space, but they do not establish universal ranking reliability. CIFAR-100 has weaker proxy--truth agreement, and DARTS is evaluated with only three searched architectures per dataset. Mass redistribution is an explicit prior for spaces with \texttt{none} and \texttt{skip}, while published search-time comparisons use different hardware and exclude final architecture training. These boundaries motivate broader search-space tests and more controlled systems comparisons.

\section{Conclusion}

We presented \method, a data-driven NAS pipeline that combines brief fair supernet training, candidate-conditioned hard-path updates, and evolutionary search. It completes NAS-Bench-201/CIFAR-10 search in 409 seconds and transfers without search-hyperparameter retuning to other datasets and the DARTS space. The central finding is that a few balanced, task-dependent updates substantially improve the architecture returned by a low-cost search, even when the best architecture visited changes little.

\subsection*{Ethics statement}

This work uses public image-classification benchmarks and involves no human
subjects or private data.  Lowering NAS cost may reduce financial and energy
barriers to architecture experimentation.  To avoid overstating this benefit,
we report hardware, wall-clock time, optimizer depth, and the exclusion of final
training from search cost.

\subsection*{AI use statement}
Generative AI tools assisted with manuscript editing, feedback on
experimental design, interpretation of results, and preparation of
appendix text and tables from experimental records. The authors take
responsibility for the final manuscript, results, and conclusions.
The authors reviewed the AI-assisted text and checked reported numerical
results against the underlying experimental records.

\subsection*{Reproducibility statement}

The released implementation records complete resolved configurations,
purpose-keyed data schedules, proposal streams, decoded genotypes, candidate
scores, winner states, timing boundaries, and artifact checksums.  A
behavioral reproduction gate confirmed exact agreement of all 780 candidate
records between the generalized implementation and the historical 60-epoch
runner, including zero numerical difference in proxy values and identical
Borda selection.  Seeds and additional configuration details are listed in
Appendix~\ref{app:config}.

\subsubsection*{Acknowledgments}

Aviral Gandhi was supported during his graduate studies by the Japanese
Government (MEXT) Scholarship. The authors thank Professor Keiki Takadama
for providing computational resources that supported this work.

\IfFileExists{iclr2027_conference.bst}{%
  \bibliographystyle{iclr2027_conference}%
}{%
  \bibliographystyle{plainnat}%
}
\bibliography{references}
\input{appendix}

\end{document}

%% file: appendix.tex
\providecommand{\method}{\textsc{LESS}\xspace}
\providecommand{\cma}{\textsc{CMA-ES}\xspace}
\appendix
\clearpage
\section{Experimental Scope, Cohorts, and Reporting Conventions}
\label{app:scope}

This appendix gives the implementation details, seed-level outcomes, additional
controls, and reproducibility information for \method. Accuracy is expressed in
percent and accuracy differences in percentage points. Unless a table says
otherwise, uncertainty is the sample standard deviation across architecture
seeds, not the uncertainty across individual validation images or candidate
evaluations. The unit of replication is an independently seeded search.

We distinguish three kinds of evidence. The principal NAS-Bench-201 (NB201)
comparison contains an initial 16-seed cohort and a subsequent eight-seed
prospective replication, each with paired $K=0$ and $K=6$ searches. Their pooled
24-seed result is a descriptive synthesis. Component ablations, dense horizon
curves, and transfer comparisons use a fixed eight-seed subset, denoted
$\mathcal S_8$. These experiments reuse some of the principal cohort's runs;
they are not additional independent replications of the headline result.
Finally, earlier development cohorts and the alternative $\sigma=0.45$ study
are reported separately in Appendix~\ref{app:development}. They are not pooled
with the frozen $\sigma=0.30$ results.

For NB201, the architecture returned by search is determined entirely by
recorded shared-weight validation losses. Benchmark training-from-scratch
accuracies are joined afterward. Benchmark validation accuracy is the primary
analysis endpoint; benchmark test accuracy is descriptive. The search's
held-out minibatch loss and the benchmark's validation accuracy are different
quantities. For DARTS, each selected genotype is trained from scratch for 600 epochs.
The primary reported endpoint is the highest test accuracy observed during
this training trajectory. Final-epoch test accuracy is reported separately
for completeness.

\subsection{Seed inventory}
\label{app:seeds}
\begingroup\scriptsize\setlength{\tabcolsep}{3pt}
\begin{longtable}{@{}p{.27\linewidth}p{.68\linewidth}@{}}
\caption{Architecture and retraining seed inventory.}\label{tab:app-seeds}\\
\toprule
Cohort & Seeds \\
\midrule\endfirsthead
\multicolumn{2}{l}{\textit{Table \thetable{} continued.}}\\
\toprule
Cohort & Seeds \\
\midrule\endhead
\midrule\multicolumn{2}{r}{\textit{Continued on next page.}}\\\endfoot
\bottomrule\endlastfoot
Initial 16 & 1333636, 2621901, 3011368, 3046636, 3274845, 3458191, 3750948, 4096113, 4564271, 4611954, 5197346, 7848893, 7990734, 8224849, 8629488, 8682075 \\
Prospective 8 & 3297086, 4007814, 4450079, 5392407, 5805960, 6963249, 8147065, 8167106 \\
Component / transfer subset & 4611954, 4564271, 2621901, 1333636, 3046636, 3274845, 4096113, 7848893 \\
DARTS architecture seeds & 184271, 392113, 618407 \\
DARTS retraining seed & 123 \\
\end{longtable}\endgroup

The eight component seeds occur within the initial 16-seed cohort. The same
$\mathcal S_8$ is used for the final CIFAR-100 depth study, ImageNet16-120
transfer, the longer-search study, and the candidate-independent calibration
control. This pairing controls architecture-seed variation within a comparison;
it does not make outcomes on different datasets statistically interchangeable.
The six DARTS searches use three architecture seeds on each dataset. All six
final retrainings use training seed 123, so their variability measures
architecture/search variation under one fixed retraining seed, not a fully
crossed search-seed-by-training-seed experiment.

\section{Complete Search Configuration and Data Protocol}
\label{app:config}

\subsection{Frozen hyperparameters}
\begingroup\scriptsize\setlength{\tabcolsep}{3pt}
\begin{longtable}{@{}p{.42\linewidth}p{.25\linewidth}p{.25\linewidth}@{}}
\caption{Frozen configuration. DARTS-specific topology is detailed in Appendix~\ref{app:darts-adapter}.}\label{tab:app-config}\\
\toprule
Setting & NB201 & DARTS \\
\midrule\endfirsthead
\multicolumn{3}{l}{\textit{Table \thetable{} continued.}}\\
\toprule
Setting & NB201 & DARTS \\
\midrule\endhead
\midrule\multicolumn{3}{r}{\textit{Continued on next page.}}\\\endfoot
\bottomrule\endlastfoot
Initial channels / cells & 16 / 8 & 16 / 8 \\
Latent logits & 30 & 224 \\
Warm-up epochs / batch & 25 / 256 & 25 / 256 \\
Warm-up LR / momentum & 0.1 cosine / 0.9 & 0.1 cosine / 0.9 \\
Warm-up weight decay / clipping & 0.0003 / 5 & 0.0003 / 5 \\
Hard paths per warm-up update & 5 & 7 \\
Search epochs / CMA distributions & 60 / 1 & 60 / 1 \\
Proposals / incumbent & 12 / 1 & 12 / 1 \\
Initial CMA sigma / mean perturbation & 0.30 / 0.45 & 0.30 / 0.45 \\
Deduplication draw cap & 50,000 & 50,000 \\
Calibration updates / group width & 6 / 5 & 6 / 7 \\
Calibration LR / momentum & 0.056 / 0.9 & 0.056 / 0.9 \\
Calibration weight decay / clipping & 0 / none & 0 / none \\
Complement schedule & candidate-specific & candidate-specific \\
Calibration batches / batch size & 6 / 256 & 6 / 256 \\
BN batches / batch size & 8 / 256 & 8 / 256 \\
Scoring batches / batch size & 8 / 128 & 8 / 128 \\
None / skip redistribution & 0.55 / 0.50 & 0.55 / 0.50 \\
CMA diversity penalty & 0 & 0 \\
Final selector & all-epoch Borda & all-epoch Borda \\
Persistent state & winner, before BN refresh & winner, before BN refresh \\
Data-loader workers & 8 & 8 \\
\end{longtable}\endgroup

The effective calibration learning rate is $0.014\times4=0.056$. Calibration
uses a new SGD optimizer for each candidate: momentum is retained across its
six local updates but never transferred to another candidate or the next epoch.
There is no calibration weight decay and no calibration gradient clipping.
Warm-up uses a separate SGD optimizer with weight decay and norm clipping.
The diversity-penalty coefficient is zero. The initial CMA scale of 0.30 is
distinct from the 0.45 standard deviation used to perturb the initial latent
mean; CMA subsequently adapts its own mean, covariance, and scale.
The initial latent center adds independent Gaussian noise of standard
deviation 0.45 to a small $10^{-3}$-scale Gaussian base encoding. In NB201
the base is the supernet's untrained architecture tensor; DARTS constructs
the corresponding array outside autograd. No benchmark accuracy enters
this initialization.

The NB201 network has 16 initial channels and eight cells. Cells at the two
stage transitions are fixed stride-two residual blocks that double the channel
count; the other cells use the searched four-node NB201 topology. The genotype
is shared across searchable cells. A convolutional stem, final normalization
and activation, global average pooling, and a classifier complete the network.
The DARTS search network also uses 16 initial channels and eight cells, with
normal and reduction genotypes and reductions at cell indices 2 and 5.

\subsection{Splits and transforms}
\begingroup\scriptsize\setlength{\tabcolsep}{3pt}
\begin{longtable}{@{}lrrrr@{}}
\caption{Search data split sizes; official test data do not score search candidates.}\label{tab:app-data}\\
\toprule
Dataset & Input & Classes & Search train & Search valid \\
\midrule\endfirsthead
\multicolumn{5}{l}{\textit{Table \thetable{} continued.}}\\
\toprule
Dataset & Input & Classes & Search train & Search valid \\
\midrule\endhead
\midrule\multicolumn{5}{r}{\textit{Continued on next page.}}\\\endfoot
\bottomrule\endlastfoot
CIFAR-10 & 32 by 32 & 10 & 25,000 & 25,000 \\
CIFAR-100 & 32 by 32 & 100 & 25,000 & 25,000 \\
ImageNet16-120 & 16 by 16 & 120 & 151,700 & 3,000 \\
\end{longtable}\endgroup

For CIFAR search, a seeded NumPy \texttt{RandomState} shuffles the 50,000
official training examples. The first 25,000 support warm-up,
calibration, and BN refresh; the remaining 25,000 supply architecture-validation
panels. The split is shared within each paired experiment. CIFAR augmentation
is a random $32\times32$ crop after four-pixel padding and random horizontal
flip, followed by conversion to a tensor and channel normalization. Validation
uses only tensor conversion and normalization. Search does not use cutout.

ImageNet16-120 uses the canonical 151,700-example training split and the
3,000-example NB201 \texttt{x-valid} split. The separate \texttt{x-test} split
is not loaded by search. Training augmentation is horizontal flip followed by
a random $16\times16$ crop with two-pixel padding. Its channel means are
$(122.68,116.66,104.01)/255$, and standard deviations are
$(63.22,61.26,65.09)/255$.

\begingroup\scriptsize\setlength{\tabcolsep}{3pt}
\begin{longtable}{@{}lp{.39\linewidth}p{.39\linewidth}@{}}
\caption{CIFAR channel normalization used by the search and DARTS retraining.}\label{tab:app-normalization}\\
\toprule
Dataset & Mean (R,G,B) & Standard deviation (R,G,B) \\
\midrule\endfirsthead
\multicolumn{3}{l}{\textit{Table \thetable{} continued.}}\\
\toprule
Dataset & Mean (R,G,B) & Standard deviation (R,G,B) \\
\midrule\endhead
\midrule\multicolumn{3}{r}{\textit{Continued on next page.}}\\\endfoot
\bottomrule\endlastfoot
CIFAR-10 & 0.49139968, 0.48215827, 0.44653124 & 0.24703233, 0.24348505, 0.26158768 \\
CIFAR-100 & 0.50707516, 0.48654887, 0.44091784 & 0.26733429, 0.25643846, 0.27615047 \\
\end{longtable}\endgroup

Warm-up iterates over the complete search-training split, retaining the final
partial minibatch. There are 98 optimizer updates per epoch for a 25,000-image
CIFAR split and 593 for ImageNet16-120. The resulting W25 budgets are 2,450 and
14,825 updates, respectively. For a warm-up horizon $W$, the learning rate used
at one-based epoch $e$ is
\begin{equation}
 \eta_e=\frac{0.1}{2}\left[1+\cos\left(\frac{\pi(e-1)}{W}\right)\right].
 \label{eq:app-warmup}
\end{equation}
Consequently, the W5, W10, W15, W25, W35, and W50 experiments train distinct
cosine schedules. They are not checkpoint prefixes of one W50 trajectory.

\subsection{Purpose-keyed data schedules}
\label{app:schedule}
The v4 scheduler keys every batch by architecture seed, epoch, purpose, batch
index, and role (indices or augmentation). Calibration, BN refresh, and scoring
have independent deterministic streams. Indices do not repeat within an
epoch's panel for a given purpose; calibration and BN panels may overlap each
other because they are independently sampled from the same training pool.
All 13 siblings use the same materialized tensors for each panel, including
the realized augmentations. Increasing $K$ extends the calibration prefix
without moving BN or scoring data. Increasing the declared search horizon
preserves earlier epochs' panels.

At $K=6$, an epoch contains six calibration batches of 256 images, eight BN
batches of 256, and eight scoring batches of 128. Thus each candidate is scored
on 1,024 images. The 1,536 calibration examples are reused across siblings
and across paths within a fair-group update; they do not represent 13
independent training panels. BN refresh has no backward pass.

\section{Implementation Details of Decoding, Calibration, and Selection}
\label{app:implementation}

\subsection{NB201 latent variables and hard execution}
NB201 has six directed edges in the order
$(0,1),(0,2),(1,2),(0,3),(1,3),(2,3)$ and five operations per edge:
\texttt{none}, \texttt{skip\_connect}, \texttt{nor\_conv\_1x1},
\texttt{nor\_conv\_3x3}, and \texttt{avg\_pool\_3x3}. Its latent dimension is
$6\times5=30$, and its discrete search space contains $5^6=15{,}625$ cells.
Decoding applies the softmax and the main-text mass redistribution, then
chooses the largest corrected
probability independently on each edge; exact operation ties follow primitive
order. The redistribution is a decoding prior, not a weighting of operations
during hard-path evaluation. A hard forward executes only the selected
operation on each edge and sums incoming edge outputs at each node.

The warm-up configuration retains historical redistribution fields
0.45/0.20. Warm-up samples hard paths directly and does not decode softmax
logits, so those fields do not change its path distribution or training
objective. Search explicitly uses 0.55/0.50. Architecture variables are not
optimized by backpropagation during either warm-up or calibration.

\subsection{Balanced NB201 gradients and state semantics}
For each calibration update, the focal operation is placed in the first row
of a five-by-six path matrix. Each edge's four remaining operations are
independently permuted into the other rows. The focal candidate is therefore
one complete hard path, and every operation occurs once on every edge.
Each path evaluates the same minibatch; dividing each cross-entropy loss by
five before backward accumulates the mean gradient. The weights are stepped
only after all five backward passes. Warm-up uses unconstrained per-edge
permutations with the same mean-gradient convention.

For clarity, write a supernet state as $(w,b)$, separating trainable weights
and BN buffers. If $v^0=0$ is the candidate's fresh momentum buffer, its
calibration weight update can be written
\begin{align}
 d^k&=\frac{1}{5}\sum_{h\in\mathcal G_{t,k}}
       \nabla_w\mathcal L_{\mathrm{tr}}(h;w^{k-1},\mathcal B^{\mathrm{cal}}_{t,k}),\\
 v^k&=0.9v^{k-1}+d^k,\qquad w^k=w^{k-1}-0.056v^k.
 \label{eq:app-momentum}
\end{align}
BN buffers evolve through the sequential training-mode forwards, rather than
through this gradient update. The checkpoint eligible for inheritance contains
the calibrated weights and those pre-refresh buffers. A separate scoring
state resets BN running statistics, uses cumulative averaging
(\texttt{momentum=None}) on eight focal-path forward passes, and then switches
to evaluation mode. Validation loss is the sum of cross-entropies divided by
the number of scored examples. The saved inheritance checkpoint is never
overwritten by this refresh. At $K=0$, calibration is absent but the same BN
refresh and scoring procedure remains.

The complement RNG is keyed by search seed, epoch, and candidate slot.
Consequently, two occurrences of the same genotype in a fallback epoch can
receive different complement groups and losses. They remain separate candidate
evaluations until the final within-epoch genotype collapse.

\subsection{CMA updates, incumbent state, and bounded deduplication}
The persistent incumbent vector and the CMA distribution mean are separate
objects. The incumbent participates in candidate evaluation and can win the
next parent state, but only the 12 proposal vectors and their losses enter
\texttt{CMA.tell}. The best recombined CMA mean is not substituted for the
lowest-loss evaluated candidate. Equal candidate losses are resolved by
evaluation order, with the incumbent evaluated first.

The decoder rejects proposals matching the incumbent or an already accepted
genotype within the epoch. At the 50,000-draw cap, all unique accepted
proposals are retained and remaining slots are filled from genuine previously
drawn duplicates. Earliest non-incumbent duplicates take priority; incumbent
duplicates are used only if needed. There are no synthesized genotypes and
no additional CMA draws after cap exhaustion. Once admitted by fallback,
these vectors are evaluated and passed to CMA like other proposals. Only
draws excluded from the final population are omitted from scoring and CMA
adaptation. Fallback epochs and filled slots are separately logged.

\subsection{Exact final selector}
\label{app:borda-exact}
Let $\bar\ell_t(g)$ be the mean loss of occurrences of genotype $g$ in epoch
$t$, and let $m_t$ be the number of distinct genotypes that epoch. The code
orders genotypes by $(\bar\ell_t(g),\operatorname{string}(g))$ and accumulates
the normalized Borda reward
\begin{equation}
 B(g)=\sum_{t:g\in\mathcal U_t} b_t(g),\qquad
 b_t(g)=\begin{cases}
 1,&m_t=1,\\
 (m_t-r_t(g))/(m_t-1),&m_t>1.
 \end{cases}
 \label{eq:app-borda-exact}
\end{equation}
The selected genotype maximizes $B(g)$; ties favor more epoch appearances,
then a later last appearance, then lexicographically smaller genotype strings.
Absent genotypes receive no contribution. With 13 distinct genotypes in every epoch, normalization multiplies
all cumulative scores by the same factor of $1/12$ relative to
unnormalized Borda scoring. The normalized rule also specifies the
contribution of epochs in which fallback reduces $m_t$.

\subsection{Computational accounting}
\begingroup\scriptsize\setlength{\tabcolsep}{3pt}
\begin{longtable}{@{}lrr@{}}
\caption{Default search workload, excluding warm-up and retraining.}\label{tab:app-counts}\\
\toprule
Quantity & NB201 & DARTS \\
\midrule\endfirsthead
\multicolumn{3}{l}{\textit{Table \thetable{} continued.}}\\
\toprule
Quantity & NB201 & DARTS \\
\midrule\endhead
\midrule\multicolumn{3}{r}{\textit{Continued on next page.}}\\\endfoot
\bottomrule\endlastfoot
Candidate evaluations & 780 & 780 \\
Calibration optimizer steps & 4680 & 4680 \\
Calibration forward/backward passes & 23400 & 32760 \\
Focal BN-refresh forwards & 6240 & 6240 \\
Validation forwards & 6240 & 6240 \\
Validation example-evaluations & 798,720 & 798,720 \\
\end{longtable}\endgroup

Search-time counts exclude warm-up and final training from scratch. The
default NB201 search uses 4,680 calibration optimizer steps and 23,400
calibration hard-path forward/backward passes. DARTS uses the same number of
optimizer steps but 32,760 passes because its fair groups have seven paths.
Both use 6,240 focal BN-refresh forwards and 6,240 validation forwards over
780 candidate evaluations. These counts explain why ``six updates'' must
not be interpreted as six single-path forward/backward passes.

\section{Statistical Definitions and Benchmark Joins}
\label{app:statistics}
For search seed $s$, let $\mathcal V_s$ contain the distinct visited
genotypes, let $g_s$ be the returned genotype, and let $A(g)$ denote
benchmark validation accuracy. We report
\begin{align}
 A_s^{\mathrm{sel}}&=A(g_s),\qquad
 A_s^{\mathrm{best}}=\max_{g\in\mathcal V_s}A(g),\\
 R_s^{\mathrm{selection}}&=A_s^{\mathrm{best}}-A_s^{\mathrm{sel}},\\
 R_s^{\mathrm{oracle}}&=A^{\mathrm{oracle}}-A_s^{\mathrm{sel}},\qquad
 \widetilde R_s=\frac{A^{\mathrm{oracle}}-A_s^{\mathrm{sel}}}
 {A^{\mathrm{oracle}}-\mu_{\mathrm{space}}}.
 \label{eq:app-regret}
\end{align}
Here $A^{\mathrm{oracle}}$ is the maximum across all 15,625 cells and
$\mu_{\mathrm{space}}$ is their mean. Normalized regret separates into
exploration and selection terms by adding and subtracting
$A_s^{\mathrm{best}}$ in the numerator. These are retrospective diagnostics,
not additional signals supplied to search.

The full-space percentile uses empirical midrank:
\[
100(\#\{A<A(g)\}+\tfrac12\#\{A=A(g)\})/15{,}625.
\]
P95 and P99 thresholds use linear-interpolated percentiles; a visited cell
counts as above a threshold only under a strict inequality. Top-five pool
quality is the mean validation accuracy of the five best distinct visited
cells. Unless explicitly labeled otherwise, proxy--truth Spearman and Kendall
correlations compare cumulative Borda scores with benchmark validation
accuracy over distinct visited genotypes. The uniform-archive analysis also
reports correlations with negative mean stored loss; these are different
diagnostics and should not be interchanged.

For paired conditions, the per-seed difference is computed before aggregation.
The two-sided Student-$t$ interval for a mean difference is
\begin{equation}
 \bar d\ \pm\ t_{0.975,n-1}\frac{s_d}{\sqrt n}.
 \label{eq:app-ci}
\end{equation}
The primary evidence suite and dense curves use 100,000 deterministic
seed-level bootstrap resamples; the separate $\sigma=0.45$ study uses
200,000. Bootstrap intervals are percentile intervals. The exact two-sided
sign test conditions on non-tied pairs. Wins/ties/losses in validation tables
refer to positive/zero/negative paired validation differences. Repeated
appearances of one architecture and benchmark training trials are not treated
as extra search seeds. Ablation intervals are exploratory and are not adjusted
for multiple comparisons.

Benchmark values use the 200-epoch training budget and deterministic averages
over the benchmark's available training trials, through
\texttt{get\_more\_info} with \texttt{hp="200"} and
\texttt{is\_random=False}. CIFAR-10 uses
\texttt{cifar10-valid}/\texttt{valid-accuracy} as its primary endpoint and
full-training \texttt{cifar10}/\texttt{test-accuracy} for the report-only test
column. CIFAR-100 uses its validation/test fields. ImageNet16-120 uses
\texttt{x-valid}/\texttt{x-test}; the independent transfer analysis verified
these mappings against every architecture in the local topology archive.

\section{CIFAR-10: Principal Paired Evaluation}
\label{app:c10}
\begingroup\scriptsize\setlength{\tabcolsep}{3pt}
\begin{longtable}{@{}lrrrrr@{}}
\caption{Principal CIFAR-10 outcomes. Test accuracy is report-only; $n$ counts search seeds.}\label{tab:app-main-agg}\\
\toprule
Cohort & Arm & n & Validation & Validation 95\% CI & Test \\
\midrule\endfirsthead
\multicolumn{6}{l}{\textit{Table \thetable{} continued.}}\\
\toprule
Cohort & Arm & n & Validation & Validation 95\% CI & Test \\
\midrule\endhead
\midrule\multicolumn{6}{r}{\textit{Continued on next page.}}\\\endfoot
\bottomrule\endlastfoot
Initial 16 & K6 & 16 & $89.848\pm0.565$ & $[89.547, 90.149]$ & $93.167\pm0.497$ \\
Initial 16 & K0 & 16 & $89.319\pm0.612$ & $[88.993, 89.645]$ & $92.694\pm0.471$ \\
Prospective 8 & K6 & 8 & $90.235\pm0.609$ & $[89.726, 90.745]$ & $93.233\pm0.429$ \\
Prospective 8 & K0 & 8 & $89.564\pm0.524$ & $[89.126, 90.003]$ & $92.692\pm0.366$ \\
Pooled 24 & K6 & 24 & $89.977\pm0.597$ & $[89.725, 90.229]$ & $93.189\pm0.467$ \\
Pooled 24 & K0 & 24 & $89.401\pm0.584$ & $[89.154, 89.647]$ & $92.693\pm0.431$ \\
\end{longtable}\endgroup

\begingroup\scriptsize\setlength{\tabcolsep}{3pt}
\begin{longtable}{@{}lrrrrrr@{}}
\caption{Paired validation differences, in percentage points. All component contrasts use eight seeds.}\label{tab:app-main-paired}\\
\toprule
Comparison & Mean & Median & t 95\% CI & Bootstrap 95\% CI & W/T/L & Sign p \\
\midrule\endfirsthead
\multicolumn{7}{l}{\textit{Table \thetable{} continued.}}\\
\toprule
Comparison & Mean & Median & t 95\% CI & Bootstrap 95\% CI & W/T/L & Sign p \\
\midrule\endhead
\midrule\multicolumn{7}{r}{\textit{Continued on next page.}}\\\endfoot
\bottomrule\endlastfoot
Initial 16: K6-K0 & 0.529 & 0.741 & $[0.086, 0.973]$ & $[0.135, 0.924]$ & 11/0/5 & 0.2101 \\
Prospective 8: K6-K0 & 0.671 & 0.480 & $[0.197, 1.145]$ & $[0.346, 1.071]$ & 8/0/0 & 0.0078 \\
Pooled 24: K6-K0 & 0.577 & 0.545 & $[0.262, 0.891]$ & $[0.287, 0.869]$ & 19/0/5 & 0.0066 \\
Subset: K6-K0 & 0.716 & 0.801 & $[0.293, 1.138]$ & $[0.379, 1.038]$ & 7/0/1 & 0.0703 \\
K6-K0-E180 & 0.643 & 0.840 & $[-0.190, 1.477]$ & $[-0.064, 1.218]$ & 6/0/2 & 0.2891 \\
CMA-uniform & 1.030 & 0.893 & $[0.466, 1.595]$ & $[0.646, 1.510]$ & 8/0/0 & 0.0078 \\
Prior on-off & 0.548 & 0.675 & $[-0.245, 1.340]$ & $[-0.088, 1.122]$ & 6/0/2 & 0.2891 \\
\end{longtable}\endgroup

The prospective eight-seed replication favors calibration on all eight
seeds. The initial cohort has 11 wins and five losses, so its sign test is
less decisive than its magnitude-based mean interval. The pooled gain of
0.577 points is a descriptive combination of these separately reported
cohorts, not a third independent experiment.

\begingroup\scriptsize\setlength{\tabcolsep}{3pt}
\begin{longtable}{@{}lrrrrrr@{}}
\caption{All 24 principal CIFAR-10 pairs. Test columns are descriptive.}\label{tab:app-main-seed}\\
\toprule
Cohort & Seed & K0 valid & K6 valid & Difference & K0 test & K6 test \\
\midrule\endfirsthead
\multicolumn{7}{l}{\textit{Table \thetable{} continued.}}\\
\toprule
Cohort & Seed & K0 valid & K6 valid & Difference & K0 test & K6 test \\
\midrule\endhead
\midrule\multicolumn{7}{r}{\textit{Continued on next page.}}\\\endfoot
\bottomrule\endlastfoot
Initial 16 & 1333636 & 89.355 & 89.880 & 0.525 & 92.890 & 93.375 \\
Initial 16 & 2621901 & 88.734 & 89.624 & 0.890 & 92.245 & 93.620 \\
Initial 16 & 3011368 & 88.798 & 90.085 & 1.287 & 92.155 & 93.630 \\
Initial 16 & 3046636 & 89.282 & 90.094 & 0.812 & 92.640 & 93.150 \\
Initial 16 & 3274845 & 88.432 & 90.024 & 1.592 & 92.235 & 93.425 \\
Initial 16 & 3458191 & 89.267 & 89.969 & 0.703 & 92.617 & 93.330 \\
Initial 16 & 3750948 & 90.094 & 89.244 & -0.850 & 93.150 & 92.470 \\
Initial 16 & 4096113 & 90.327 & 91.116 & 0.789 & 93.397 & 94.020 \\
Initial 16 & 4564271 & 90.094 & 89.900 & -0.194 & 93.150 & 93.315 \\
Initial 16 & 4611954 & 89.252 & 90.132 & 0.880 & 92.730 & 93.460 \\
Initial 16 & 5197346 & 89.520 & 89.052 & -0.468 & 92.815 & 92.635 \\
Initial 16 & 7848893 & 89.124 & 89.555 & 0.431 & 92.260 & 92.740 \\
Initial 16 & 7990734 & 89.274 & 90.053 & 0.779 & 92.640 & 93.210 \\
Initial 16 & 8224849 & 88.997 & 88.733 & -0.264 & 92.687 & 92.030 \\
Initial 16 & 8629488 & 88.328 & 90.520 & 2.192 & 91.893 & 93.315 \\
Initial 16 & 8682075 & 90.223 & 89.588 & -0.635 & 93.600 & 92.945 \\
Prospective 8 & 3297086 & 89.897 & 90.413 & 0.516 & 92.697 & 93.157 \\
Prospective 8 & 4007814 & 89.526 & 89.853 & 0.327 & 92.950 & 93.103 \\
Prospective 8 & 4450079 & 90.555 & 91.119 & 0.564 & 93.263 & 93.850 \\
Prospective 8 & 5392407 & 89.382 & 89.586 & 0.204 & 92.765 & 92.660 \\
Prospective 8 & 5805960 & 88.798 & 90.094 & 1.296 & 92.230 & 93.150 \\
Prospective 8 & 6963249 & 89.790 & 90.234 & 0.444 & 92.955 & 93.285 \\
Prospective 8 & 8147065 & 89.277 & 91.064 & 1.787 & 92.400 & 93.845 \\
Prospective 8 & 8167106 & 89.290 & 89.520 & 0.230 & 92.280 & 92.815 \\
\end{longtable}\endgroup

\begingroup\scriptsize\setlength{\tabcolsep}{3pt}
\begin{longtable}{@{}lrrrrrr@{}}
\caption{Pool and ranking diagnostics for the principal cohorts; entries are seed means.}\label{tab:app-main-diag}\\
\toprule
Cohort & Arm & Best visited & Regret & Unique & Spearman & Kendall \\
\midrule\endfirsthead
\multicolumn{7}{l}{\textit{Table \thetable{} continued.}}\\
\toprule
Cohort & Arm & Best visited & Regret & Unique & Spearman & Kendall \\
\midrule\endhead
\midrule\multicolumn{7}{r}{\textit{Continued on next page.}}\\\endfoot
\bottomrule\endlastfoot
Initial 16 & K6 & 90.661 & 0.812 & 127.2 & 0.508 & 0.358 \\
Initial 16 & K0 & 90.629 & 1.311 & 144.5 & 0.474 & 0.333 \\
Prospective 8 & K6 & 90.870 & 0.635 & 120.1 & 0.469 & 0.336 \\
Prospective 8 & K0 & 90.769 & 1.204 & 152.1 & 0.440 & 0.306 \\
Pooled 24 & K6 & 90.730 & 0.753 & 124.9 & 0.495 & 0.350 \\
Pooled 24 & K0 & 90.676 & 1.275 & 147.0 & 0.463 & 0.324 \\
\end{longtable}\endgroup

The pooled best-visited accuracy changes by only about 0.054 points, while
selection regret falls by about 0.523 points. The larger change is therefore
in conversion of a visited pool into the returned architecture. This
decomposition is descriptive and does not isolate a single causal component:
calibration also changes the winner's inherited state and hence future search.

\subsection{Operation composition}
\begingroup\scriptsize\setlength{\tabcolsep}{3pt}
\begin{longtable}{@{}lrrrrrr@{}}
\caption{Mean selected-cell operation counts (six edges per cell).}\label{tab:app-main-ops}\\
\toprule
Cohort & Arm & None & Skip & Conv 1 & Conv 3 & Pool \\
\midrule\endfirsthead
\multicolumn{7}{l}{\textit{Table \thetable{} continued.}}\\
\toprule
Cohort & Arm & None & Skip & Conv 1 & Conv 3 & Pool \\
\midrule\endhead
\midrule\multicolumn{7}{r}{\textit{Continued on next page.}}\\\endfoot
\bottomrule\endlastfoot
Initial 16 & K0 & 0.438 & 1.062 & 1.688 & 2.375 & 0.438 \\
Initial 16 & K6 & 0.125 & 0.562 & 2.375 & 2.875 & 0.062 \\
Prospective 8 & K0 & 0.250 & 1.375 & 1.750 & 2.125 & 0.500 \\
Prospective 8 & K6 & 0.125 & 1.000 & 2.250 & 2.625 & 0.000 \\
Pooled 24 & K0 & 0.375 & 1.167 & 1.708 & 2.292 & 0.458 \\
Pooled 24 & K6 & 0.125 & 0.708 & 2.333 & 2.792 & 0.042 \\
\end{longtable}\endgroup

Operation counts refer to the six edges of each returned NB201 cell. They
are descriptive outcomes of the frozen decoder and search, not additional
constraints imposed on returned architectures.

\section{Component Controls and Alternative Selection Rules}
\label{app:controls}
\begingroup\scriptsize\setlength{\tabcolsep}{3pt}
\begin{longtable}{@{}lrrrrrr@{}}
\caption{Matched component controls on $\mathcal S_8$. Test is report-only.}\label{tab:app-control-agg}\\
\toprule
Arm & Validation & Test & Best visited & Regret & Unique & Spearman \\
\midrule\endfirsthead
\multicolumn{7}{l}{\textit{Table \thetable{} continued.}}\\
\toprule
Arm & Validation & Test & Best visited & Regret & Unique & Spearman \\
\midrule\endhead
\midrule\multicolumn{7}{r}{\textit{Continued on next page.}}\\\endfoot
\bottomrule\endlastfoot
K6 reference & $90.041\pm0.482$ & $93.388\pm0.367$ & 90.663 & 0.623 & 132.9 & 0.525 \\
K0 E180 & $89.397\pm0.931$ & $92.762\pm0.613$ & 90.759 & 1.362 & 216.9 & 0.381 \\
Uniform K6 & $89.010\pm0.288$ & $92.078\pm0.149$ & 91.389 & 2.379 & 704.5 & 0.721 \\
No prior K6 & $89.493\pm0.837$ & $92.486\pm0.660$ & 91.103 & 1.610 & 187.2 & 0.373 \\
\end{longtable}\endgroup

\begingroup\scriptsize\setlength{\tabcolsep}{3pt}
\begin{longtable}{@{}lrrrr@{}}
\caption{Per-seed selected validation for matched component controls.}\label{tab:app-control-seed}\\
\toprule
Seed & CMA K6 & K0 E180 & Uniform K6 & Prior off K6 \\
\midrule\endfirsthead
\multicolumn{5}{l}{\textit{Table \thetable{} continued.}}\\
\toprule
Seed & CMA K6 & K0 E180 & Uniform K6 & Prior off K6 \\
\midrule\endhead
\midrule\multicolumn{5}{r}{\textit{Continued on next page.}}\\\endfoot
\bottomrule\endlastfoot
4611954 & 90.132 & 88.662 & 89.202 & 88.712 \\
4564271 & 89.900 & 90.094 & 89.005 & 89.300 \\
2621901 & 89.624 & 88.734 & 89.119 & 88.873 \\
1333636 & 89.880 & 89.355 & 89.328 & 90.694 \\
3046636 & 90.094 & 88.670 & 88.451 & 89.060 \\
3274845 & 90.024 & 88.432 & 89.132 & 90.904 \\
4096113 & 91.116 & 90.327 & 88.711 & 89.405 \\
7848893 & 89.555 & 90.904 & 89.134 & 88.995 \\
\end{longtable}\endgroup

\subsection{More evaluations without calibration}
The expanded-budget control runs $K=0$ for 180 epochs, giving 2,340 candidate
evaluations versus 780 for the default $K=6$ search. Its mean complete cost
is 529.0 seconds versus 410.2 seconds for the matched K6 subset. Thus the
realized control is an expanded-budget comparison, not an exactly equal-time
experiment: it costs about $1.29\times$ as much. The K6-minus-K0-E180
validation difference is 0.643 points with a $t$ interval spanning zero.
The separate selector duration of the continued E180 run was not recorded;
its zero-valued placeholder in the source runtime table is not evidence of
zero selection cost.

\subsection{CMA versus uniform proposals and the decoder prior}
The uniform control changes the proposal mechanism while keeping the
candidate evaluator and inherited-state protocol. Its much larger visited
archive reaches better best-visited validation accuracy but returns a lower
accuracy architecture. CMA exceeds uniform on all eight matched seeds, with
a mean validation difference of 1.030 points. Removing mass redistribution
also increases archive breadth, but the mean selected-validation effect of
keeping the prior, 0.548 points, has an interval spanning zero. These controls
support treating exploration quality and final selection quality separately.

\subsection{Replay of the final selection rule}
\begingroup\scriptsize\setlength{\tabcolsep}{3pt}
\begin{longtable}{@{}lrrrr@{}}
\caption{Recorded offline selector replay on the initial 16 K6 trajectories.}\label{tab:app-selectors}\\
\toprule
Selector & n & Validation & Median & t 95\% CI \\
\midrule\endfirsthead
\multicolumn{5}{l}{\textit{Table \thetable{} continued.}}\\
\toprule
Selector & n & Validation & Median & t 95\% CI \\
\midrule\endhead
\midrule\multicolumn{5}{r}{\textit{Continued on next page.}}\\\endfoot
\bottomrule\endlastfoot
Borda & 16 & $89.848\pm0.565$ & 89.935 & $[89.547, 90.149]$ \\
Final-epoch input incumbent & 16 & $89.881\pm0.646$ & 89.902 & $[89.537, 90.225]$ \\
Last-epoch winner & 16 & $89.955\pm0.649$ & 89.878 & $[89.609, 90.301]$ \\
\end{longtable}\endgroup

These alternatives are computed offline on the same recorded trajectories;
they do not change proposals, candidate calibration, or inherited states.
The source selector-replay table's ``final incumbent'' is the order-zero
candidate entering the final epoch, whereas ``last epoch'' is the winner
selected within that epoch. We label the former as the final-epoch input
incumbent to distinguish it from the incumbent after the last update.
The close results do not establish a Borda-specific gain. This analysis
replays stored scores without additional candidate evaluations.

\section{Candidate-Conditioned versus Candidate-Independent Calibration}
\label{app:conditioned}
This matched control replaces the focal-anchored fair group by independent
per-edge permutations of all five operations. It preserves six optimizer
steps, 30 hard-path backward passes, shared panels, the warm-up checkpoint,
and the remaining search settings. An independent group can contain the
candidate by chance; its construction does not use the focal genotype.
Thus the comparison targets candidate conditioning at a fixed update budget.

\begingroup\scriptsize\setlength{\tabcolsep}{3pt}
\begin{longtable}{@{}lrrrrrr@{}}
\caption{Candidate conditioning control on eight paired seeds; means with sample SD.}\label{tab:app-condition-agg}\\
\toprule
Arm & Validation & Test & Best visited & Regret & Unique & Time (s) \\
\midrule\endfirsthead
\multicolumn{7}{l}{\textit{Table \thetable{} continued.}}\\
\toprule
Arm & Validation & Test & Best visited & Regret & Unique & Time (s) \\
\midrule\endhead
\midrule\multicolumn{7}{r}{\textit{Continued on next page.}}\\\endfoot
\bottomrule\endlastfoot
Conditioned & $90.041\pm0.482$ & $93.388\pm0.367$ & $90.663\pm0.589$ & $0.623\pm0.358$ & $132.9\pm45.4$ & $410.2\pm2.6$ \\
Independent & $89.031\pm2.131$ & $92.451\pm1.889$ & $90.611\pm0.462$ & $1.580\pm1.909$ & $115.2\pm41.6$ & $407.5\pm3.7$ \\
\end{longtable}\endgroup

\begingroup\scriptsize\setlength{\tabcolsep}{3pt}
\begin{longtable}{@{}lrrrrrrr@{}}
\caption{Candidate-conditioned (C) versus independent (I) fair calibration, per seed.}\label{tab:app-condition-seed}\\
\toprule
Seed & C valid & I valid & C-I & C best & I best & C regret & I regret \\
\midrule\endfirsthead
\multicolumn{8}{l}{\textit{Table \thetable{} continued.}}\\
\toprule
Seed & C valid & I valid & C-I & C best & I best & C regret & I regret \\
\midrule\endhead
\midrule\multicolumn{8}{r}{\textit{Continued on next page.}}\\\endfoot
\bottomrule\endlastfoot
4611954 & 90.132 & 89.588 & 0.544 & 91.533 & 90.164 & 1.401 & 0.576 \\
4564271 & 89.900 & 89.900 & 0.000 & 90.712 & 90.712 & 0.812 & 0.812 \\
2621901 & 89.624 & 89.520 & 0.104 & 90.106 & 90.474 & 0.482 & 0.954 \\
1333636 & 89.880 & 90.024 & -0.144 & 90.386 & 90.712 & 0.506 & 0.688 \\
3046636 & 90.094 & 87.831 & 2.263 & 90.302 & 90.806 & 0.208 & 2.975 \\
3274845 & 90.024 & 90.024 & 0.000 & 90.466 & 90.386 & 0.442 & 0.362 \\
4096113 & 91.116 & 91.105 & 0.011 & 91.607 & 91.553 & 0.491 & 0.448 \\
7848893 & 89.555 & 84.260 & 5.295 & 90.194 & 90.084 & 0.639 & 5.824 \\
\end{longtable}\endgroup

\begingroup\scriptsize\setlength{\tabcolsep}{3pt}
\begin{longtable}{@{}lrrrr@{}}
\caption{Conditioned-minus-independent effects. All entries are paired; runtime is seconds.}\label{tab:app-condition-paired}\\
\toprule
Metric & Mean & Median & t 95\% CI & Bootstrap 95\% CI \\
\midrule\endfirsthead
\multicolumn{5}{l}{\textit{Table \thetable{} continued.}}\\
\toprule
Metric & Mean & Median & t 95\% CI & Bootstrap 95\% CI \\
\midrule\endhead
\midrule\multicolumn{5}{r}{\textit{Continued on next page.}}\\\endfoot
\bottomrule\endlastfoot
selected validation & 1.009 & 0.057 & $[-0.581, 2.600]$ & $[0.029, 2.349]$ \\
best visited validation & 0.052 & 0.027 & $[-0.435, 0.538]$ & $[-0.259, 0.477]$ \\
selection regret & -0.957 & -0.091 & $[-2.634, 0.719]$ & $[-2.379, 0.144]$ \\
complete method seconds & 2.636 & 3.563 & $[-0.046, 5.319]$ & $[0.533, 4.641]$ \\
\end{longtable}\endgroup

The mean conditioned-minus-independent validation effect is 1.009 points,
but the median is only 0.057 points. Two seeds account for most of the mean
difference; there are five wins, two ties, and one loss. The $t$ interval
includes zero and the exact sign-test $p$-value is 0.219, while the percentile
bootstrap interval is positive. The result is suggestive of protection
against occasional poor selections, not precise evidence of a uniform
per-seed advantage. Best-visited validation changes by only 0.052 points.

\section{Dense Calibration-Depth and Warm-Up Ablations}
\label{app:dense}
The dense CIFAR-10 study merges 56 newly completed odd-$K$/intermediate-$W$
runs with compatible previously completed settings. All points use
$\mathcal S_8$, $\sigma=0.30$, and 60 search epochs. There are 112 distinct
runs: 72 K-curve runs and 48 W-curve runs, with eight K6/W25 references shared
by both curves. The reference is counted once when describing independent
search executions. Each K arm reuses the same seed's W25 checkpoint.

\subsection{Calibration depth at W25}
\begingroup\scriptsize\setlength{\tabcolsep}{3pt}
\begin{longtable}{@{}lrrrrrrr@{}}
\caption{Dense K curve on eight CIFAR-10 seeds. Test is report-only; runtime retains all measurements.}\label{tab:app-k-agg}\\
\toprule
K & Validation & Test & Best & Regret & Unique & Spearman & Time (s) \\
\midrule\endfirsthead
\multicolumn{8}{l}{\textit{Table \thetable{} continued.}}\\
\toprule
K & Validation & Test & Best & Regret & Unique & Spearman & Time (s) \\
\midrule\endhead
\midrule\multicolumn{8}{r}{\textit{Continued on next page.}}\\\endfoot
\bottomrule\endlastfoot
0 & $89.325\pm0.631$ & $92.693\pm0.438$ & 90.679 & 1.354 & 149.1 & 0.447 & $205.4\pm5.8$ \\
1 & $90.131\pm0.769$ & $93.344\pm0.611$ & 90.961 & 0.830 & 138.4 & 0.424 & $243.1\pm3.6$ \\
2 & $89.726\pm0.761$ & $92.951\pm0.535$ & 90.648 & 0.922 & 144.6 & 0.393 & $278.4\pm3.4$ \\
3 & $89.510\pm0.668$ & $92.972\pm0.505$ & 90.488 & 0.978 & 128.8 & 0.481 & $310.5\pm1.7$ \\
4 & $89.378\pm0.367$ & $92.722\pm0.387$ & 90.543 & 1.165 & 126.9 & 0.467 & $344.0\pm2.6$ \\
5 & $89.909\pm0.598$ & $93.216\pm0.475$ & 90.520 & 0.611 & 110.1 & 0.506 & $375.8\pm2.4$ \\
6 & $90.041\pm0.482$ & $93.388\pm0.367$ & 90.663 & 0.623 & 132.9 & 0.525 & $410.2\pm2.6$ \\
7 & $89.760\pm0.232$ & $93.150\pm0.238$ & 90.683 & 0.923 & 108.1 & 0.525 & $443.9\pm5.5$ \\
8 & $89.575\pm0.297$ & $92.952\pm0.461$ & 90.470 & 0.895 & 104.4 & 0.548 & $477.0\pm3.7$ \\
\end{longtable}\endgroup

\begingroup\scriptsize\setlength{\tabcolsep}{3pt}
\begin{longtable}{@{}lrrrrrrrrr@{}}
\caption{Per-seed selected validation for the dense K curve.}\label{tab:app-k-seed}\\
\toprule
Seed & K0 & K1 & K2 & K3 & K4 & K5 & K6 & K7 & K8 \\
\midrule\endfirsthead
\multicolumn{10}{l}{\textit{Table \thetable{} continued.}}\\
\toprule
Seed & K0 & K1 & K2 & K3 & K4 & K5 & K6 & K7 & K8 \\
\midrule\endhead
\midrule\multicolumn{10}{r}{\textit{Continued on next page.}}\\\endfoot
\bottomrule\endlastfoot
4611954 & 89.252 & 89.461 & 89.461 & 89.252 & 89.461 & 89.626 & 90.132 & 89.626 & 89.461 \\
4564271 & 90.094 & 90.268 & 90.268 & 90.046 & 89.900 & 89.900 & 89.900 & 89.900 & 89.774 \\
2621901 & 88.734 & 89.624 & 88.441 & 89.520 & 89.626 & 89.520 & 89.624 & 89.488 & 89.624 \\
1333636 & 89.355 & 90.024 & 89.928 & 90.820 & 89.679 & 89.928 & 89.880 & 89.928 & 89.277 \\
3046636 & 89.282 & 90.094 & 89.592 & 89.078 & 89.136 & 89.588 & 90.094 & 89.382 & 89.591 \\
3274845 & 88.432 & 91.344 & 91.098 & 89.624 & 89.234 & 89.830 & 90.024 & 90.024 & 90.024 \\
4096113 & 90.327 & 91.098 & 89.467 & 88.735 & 88.734 & 91.334 & 91.116 & 89.853 & 89.086 \\
7848893 & 89.124 & 89.134 & 89.555 & 89.006 & 89.252 & 89.546 & 89.555 & 89.880 & 89.762 \\
\end{longtable}\endgroup

\begingroup\scriptsize\setlength{\tabcolsep}{3pt}
\begin{longtable}{@{}lrrrrr@{}}
\caption{Dense K comparisons against the frozen reference. No multiplicity correction.}\label{tab:app-k-paired}\\
\toprule
Comparison & Mean & Median & t 95\% CI & Bootstrap 95\% CI & W/T/L \\
\midrule\endfirsthead
\multicolumn{6}{l}{\textit{Table \thetable{} continued.}}\\
\toprule
Comparison & Mean & Median & t 95\% CI & Bootstrap 95\% CI & W/T/L \\
\midrule\endhead
\midrule\multicolumn{6}{r}{\textit{Continued on next page.}}\\\endfoot
\bottomrule\endlastfoot
K0 - K6 & -0.716 & -0.801 & $[-1.138, -0.293]$ & $[-1.039, -0.382]$ & 1/0/7 \\
K1 - K6 & 0.090 & 0.000 & $[-0.405, 0.586]$ & $[-0.257, 0.504]$ & 3/2/3 \\
K2 - K6 & -0.314 & -0.251 & $[-1.043, 0.415]$ & $[-0.871, 0.244]$ & 3/1/4 \\
K3 - K6 & -0.530 & -0.474 & $[-1.342, 0.281]$ & $[-1.189, 0.071]$ & 2/0/6 \\
K4 - K6 & -0.663 & -0.487 & $[-1.317, -0.009]$ & $[-1.221, -0.232]$ & 1/1/6 \\
K5 - K6 & -0.132 & -0.056 & $[-0.349, 0.085]$ & $[-0.304, 0.029]$ & 2/1/5 \\
K7 - K6 & -0.280 & -0.068 & $[-0.711, 0.150]$ & $[-0.637, 0.025]$ & 2/2/4 \\
K8 - K6 & -0.466 & -0.315 & $[-1.058, 0.126]$ & $[-0.978, -0.079]$ & 1/2/5 \\
\end{longtable}\endgroup

All nonzero K settings have a higher mean selected validation accuracy than
K0 on these seeds, but the curve is not monotonic. K1 has a slightly higher
mean than K6 and a much lower cost; its paired comparison with K6 remains
imprecise. The frozen choice remains K6. This eight-seed sensitivity analysis does not establish a unique
optimal calibration depth. Greater pool-wide rank correlation at
larger K does not guarantee a better top-ranked returned architecture.

\subsection{Warm-up horizon at K6}
\begingroup\scriptsize\setlength{\tabcolsep}{3pt}
\begin{longtable}{@{}lrrrrrrr@{}}
\caption{Dense W curve on eight CIFAR-10 seeds. Test is report-only; runtime retains all measurements.}\label{tab:app-w-agg}\\
\toprule
W & Validation & Test & Best & Regret & Unique & Spearman & Time (s) \\
\midrule\endfirsthead
\multicolumn{8}{l}{\textit{Table \thetable{} continued.}}\\
\toprule
W & Validation & Test & Best & Regret & Unique & Spearman & Time (s) \\
\midrule\endhead
\midrule\multicolumn{8}{r}{\textit{Continued on next page.}}\\\endfoot
\bottomrule\endlastfoot
5 & $88.865\pm1.817$ & $92.253\pm1.853$ & 90.560 & 1.695 & 114.2 & 0.187 & $327.8\pm6.0$ \\
10 & $89.715\pm0.423$ & $93.066\pm0.408$ & 90.805 & 1.090 & 134.6 & 0.356 & $348.9\pm2.8$ \\
15 & $89.762\pm0.360$ & $93.114\pm0.400$ & 90.451 & 0.689 & 129.2 & 0.455 & $368.9\pm2.9$ \\
25 & $90.041\pm0.482$ & $93.388\pm0.367$ & 90.663 & 0.623 & 132.9 & 0.525 & $410.2\pm2.6$ \\
35 & $89.837\pm0.809$ & $92.969\pm0.553$ & 90.759 & 0.923 & 122.4 & 0.494 & $447.5\pm2.1$ \\
50 & $89.880\pm0.289$ & $93.168\pm0.252$ & 90.634 & 0.754 & 136.9 & 0.507 & $509.3\pm1.9$ \\
\end{longtable}\endgroup

\begingroup\scriptsize\setlength{\tabcolsep}{3pt}
\begin{longtable}{@{}lrrrrrr@{}}
\caption{Per-seed selected validation for the dense W curve.}\label{tab:app-w-seed}\\
\toprule
Seed & W5 & W10 & W15 & W25 & W35 & W50 \\
\midrule\endfirsthead
\multicolumn{7}{l}{\textit{Table \thetable{} continued.}}\\
\toprule
Seed & W5 & W10 & W15 & W25 & W35 & W50 \\
\midrule\endhead
\midrule\multicolumn{7}{r}{\textit{Continued on next page.}}\\\endfoot
\bottomrule\endlastfoot
4611954 & 86.264 & 89.461 & 89.461 & 90.132 & 89.586 & 89.461 \\
4564271 & 89.508 & 89.774 & 89.900 & 89.900 & 89.900 & 89.900 \\
2621901 & 89.744 & 89.624 & 89.006 & 89.624 & 89.626 & 89.944 \\
1333636 & 90.024 & 88.835 & 90.024 & 89.880 & 90.477 & 89.880 \\
3046636 & 85.672 & 90.108 & 89.948 & 90.094 & 88.659 & 90.094 \\
3274845 & 89.728 & 89.800 & 90.024 & 90.024 & 90.108 & 89.488 \\
4096113 & 90.412 & 90.119 & 89.996 & 91.116 & 91.260 & 89.944 \\
7848893 & 89.568 & 89.996 & 89.735 & 89.555 & 89.078 & 90.331 \\
\end{longtable}\endgroup

\begingroup\scriptsize\setlength{\tabcolsep}{3pt}
\begin{longtable}{@{}lrrrrr@{}}
\caption{Dense W comparisons against the frozen reference. No multiplicity correction.}\label{tab:app-w-paired}\\
\toprule
Comparison & Mean & Median & t 95\% CI & Bootstrap 95\% CI & W/T/L \\
\midrule\endfirsthead
\multicolumn{6}{l}{\textit{Table \thetable{} continued.}}\\
\toprule
Comparison & Mean & Median & t 95\% CI & Bootstrap 95\% CI & W/T/L \\
\midrule\endhead
\midrule\multicolumn{6}{r}{\textit{Continued on next page.}}\\\endfoot
\bottomrule\endlastfoot
W5 - W25 & -1.176 & -0.344 & $[-2.731, 0.380]$ & $[-2.507, -0.109]$ & 3/0/5 \\
W10 - W25 & -0.326 & -0.175 & $[-0.767, 0.115]$ & $[-0.671, 0.011]$ & 2/1/5 \\
W15 - W25 & -0.279 & -0.073 & $[-0.671, 0.113]$ & $[-0.598, 0.002]$ & 2/2/4 \\
W35 - W25 & -0.204 & 0.001 & $[-0.717, 0.309]$ & $[-0.637, 0.157]$ & 4/1/3 \\
W50 - W25 & -0.160 & 0.000 & $[-0.671, 0.350]$ & $[-0.567, 0.230]$ & 2/3/3 \\
\end{longtable}\endgroup

Shortening warm-up to five epochs increases seed-to-seed variability and
selection regret. Increasing warm-up beyond 25 epochs increases cost without
a resolved mean improvement on these eight seeds. Every W setting has its
own cosine schedule; the comparison jointly changes training duration and
the learning-rate path.

The merged analysis retains all runs, including two K0 searches with roughly
uniform wall-clock slowdowns. It reports the full-sample K0 mean and the
204.8-second median. The evidence-suite runtime table instead excludes
those two measurements only from its timing summaries; all accuracy
statistics retain them. The dense tables here follow the merged convention.
One fallback slot occurred at K7, seed 4564271, epoch 59; the other 111 runs
used strict deduplication throughout. No candidate failures or non-finite
records were reported in the 87,360 candidate events.

\section{Longer Search and Late-Trajectory Concentration}
\label{app:budget}
\begingroup\scriptsize\setlength{\tabcolsep}{3pt}
\begin{longtable}{@{}lrrrrrr@{}}
\caption{Borda selection on prefixes of eight E120 trajectories; means with sample SD.}\label{tab:app-budget-agg}\\
\toprule
Prefix & Validation & Test & Best & Regret & Unique & Time (s) \\
\midrule\endfirsthead
\multicolumn{7}{l}{\textit{Table \thetable{} continued.}}\\
\toprule
Prefix & Validation & Test & Best & Regret & Unique & Time (s) \\
\midrule\endhead
\midrule\multicolumn{7}{r}{\textit{Continued on next page.}}\\\endfoot
\bottomrule\endlastfoot
E30 & $89.976\pm0.514$ & $93.319\pm0.386$ & $90.655\pm0.596$ & $0.679\pm0.538$ & $95.5\pm30.5$ & $257.3\pm0.9$ \\
E60 & $90.041\pm0.482$ & $93.388\pm0.367$ & $90.663\pm0.589$ & $0.623\pm0.358$ & $132.9\pm45.4$ & $410.3\pm2.2$ \\
E90 & $90.203\pm0.550$ & $93.519\pm0.303$ & $90.729\pm0.623$ & $0.525\pm0.352$ & $145.6\pm55.8$ & $567.4\pm8.1$ \\
E120 & $90.183\pm0.587$ & $93.432\pm0.391$ & $90.765\pm0.596$ & $0.582\pm0.359$ & $153.5\pm57.8$ & $731.4\pm37.6$ \\
\end{longtable}\endgroup

\begingroup\scriptsize\setlength{\tabcolsep}{3pt}
\begin{longtable}{@{}lrrrrrrrr@{}}
\caption{Per-seed selected validation and unique-genotype counts at each prefix.}\label{tab:app-budget-seed}\\
\toprule
Seed & E30 val & E60 val & E90 val & E120 val & E30 uniq & E60 uniq & E90 uniq & E120 uniq \\
\midrule\endfirsthead
\multicolumn{9}{l}{\textit{Table \thetable{} continued.}}\\
\toprule
Seed & E30 val & E60 val & E90 val & E120 val & E30 uniq & E60 uniq & E90 uniq & E120 uniq \\
\midrule\endhead
\midrule\multicolumn{9}{r}{\textit{Continued on next page.}}\\\endfoot
\bottomrule\endlastfoot
4611954 & 89.588 & 90.132 & 91.012 & 91.012 & 115 & 161 & 167 & 171 \\
4564271 & 89.900 & 89.900 & 89.900 & 89.900 & 68 & 87 & 103 & 108 \\
2621901 & 89.624 & 89.624 & 89.624 & 89.626 & 88 & 104 & 109 & 120 \\
1333636 & 89.880 & 89.880 & 89.880 & 89.526 & 99 & 133 & 136 & 136 \\
3046636 & 90.094 & 90.094 & 90.094 & 90.094 & 74 & 97 & 98 & 104 \\
3274845 & 90.024 & 90.024 & 90.024 & 90.024 & 97 & 131 & 154 & 178 \\
4096113 & 91.146 & 91.116 & 91.116 & 91.116 & 158 & 229 & 270 & 280 \\
7848893 & 89.555 & 89.555 & 89.978 & 90.164 & 65 & 121 & 128 & 131 \\
\end{longtable}\endgroup

\begingroup\scriptsize\setlength{\tabcolsep}{3pt}
\begin{longtable}{@{}lrrrr@{}}
\caption{Selected-validation differences between trajectory prefixes.}\label{tab:app-budget-paired}\\
\toprule
Contrast & Mean & t 95\% CI & Bootstrap 95\% CI & W/T/L \\
\midrule\endfirsthead
\multicolumn{5}{l}{\textit{Table \thetable{} continued.}}\\
\toprule
Contrast & Mean & t 95\% CI & Bootstrap 95\% CI & W/T/L \\
\midrule\endhead
\midrule\multicolumn{5}{r}{\textit{Continued on next page.}}\\\endfoot
\bottomrule\endlastfoot
E120-E60 & 0.142 & [-0.191, 0.475] & [-0.088, 0.407] & 3/4/1 \\
E90-E60 & 0.163 & [-0.109, 0.435] & [0.000, 0.383] & 2/6/0 \\
E120-E90 & -0.021 & [-0.146, 0.104] & [-0.133, 0.070] & 2/5/1 \\
E60-E30 & 0.064 & [-0.098, 0.227] & [-0.011, 0.204] & 1/6/1 \\
\end{longtable}\endgroup

\begingroup\scriptsize\setlength{\tabcolsep}{3pt}
\begin{longtable}{@{}lrrrrrr@{}}
\caption{Mean 30-epoch block diagnostics over eight seeds; fallback is mean events per seed.}\label{tab:app-budget-blocks}\\
\toprule
Block & Raw draws & Acceptance & Duplicate rej. & Incumbent rej. & New cells & Fallbacks \\
\midrule\endfirsthead
\multicolumn{7}{l}{\textit{Table \thetable{} continued.}}\\
\toprule
Block & Raw draws & Acceptance & Duplicate rej. & Incumbent rej. & New cells & Fallbacks \\
\midrule\endhead
\midrule\multicolumn{7}{r}{\textit{Continued on next page.}}\\\endfoot
\bottomrule\endlastfoot
1--30 & 2,145.8 & 0.325 & 0.499 & 0.176 & 95.5 & 0.000 \\
31--60 & 2,689.0 & 0.230 & 0.563 & 0.207 & 37.4 & 0.000 \\
61--90 & 31,018.2 & 0.076 & 0.719 & 0.206 & 12.8 & 0.125 \\
91--120 & 87,594.6 & 0.139 & 0.652 & 0.210 & 7.9 & 1.500 \\
\end{longtable}\endgroup

The longer study runs each seed to 120 epochs and replays Borda over prefixes
of 30, 60, 90, and 120 epochs. These are correlated prefixes of eight
trajectories, not 32 independent searches. From E60 to E120, the mean gain
in selected validation is 0.142 points and the median gain is zero. Four
seeds tie, three improve, and one declines. The interval includes zero,
while the additional complete cost is about 321 seconds. New-genotype
discovery becomes sparse late in the trajectory and raw proposal rejection
increases. Initial $\sigma=0.30$ is known, but these NB201 logs do not expose
the full adaptive internal CMA-scale trajectory; we do not reconstruct it.

\section{Uniform-Archive Size and Repeated-Evaluation Analyses}
\label{app:archive}
The offline archive study uses the eight completed CMA and uniform K6
trajectories. Protocol A keeps the shortest whole-epoch uniform prefix whose
unique-genotype count reaches the matched CMA count. Protocol B samples
exactly that many unique uniform genotypes and retains their stored
occurrences, repeating the sampling 200 times per seed. Both recompute
within-epoch rankings from stored losses and replay the same Borda rule;
neither performs a new network evaluation.

\begingroup\scriptsize\setlength{\tabcolsep}{3pt}
\begin{longtable}{@{}lrrrr@{}}
\caption{Offline archive-size controls. Protocol B entries average 200 subsamples within each seed.}\label{tab:app-archive}\\
\toprule
Quantity & CMA & Uniform & A: prefix & B: subsample \\
\midrule\endfirsthead
\multicolumn{5}{l}{\textit{Table \thetable{} continued.}}\\
\toprule
Quantity & CMA & Uniform & A: prefix & B: subsample \\
\midrule\endhead
\midrule\multicolumn{5}{r}{\textit{Continued on next page.}}\\\endfoot
\bottomrule\endlastfoot
Unique genotypes & 132.9 & 704.5 & 139.6 & 132.9 \\
Retained epoch-evaluations & 780 & 780 & 151.1 & 147.1 \\
Epochs per genotype & 6.38 & 1.11 & 1.08 & 1.11 \\
Selected validation & 90.04 & 89.01 & 89.35 & 88.92 \\
Selected validation t CI & [89.64,90.44] & [88.77,89.25] & [88.79,89.90] & [88.50,89.33] \\
Test (report-only) & 93.39 & 92.08 & 92.55 & 92.06 \\
Best visited & 90.66 & 91.39 & 91.06 & 91.08 \\
Within-archive regret & 0.62 & 2.38 & 1.71 & 2.16 \\
Mean archive validation & 89.10 & 83.51 & 83.19 & 83.51 \\
Spearman, negative mean loss & 0.54 & 0.72 & 0.69 & 0.71 \\
Spearman, Borda & 0.52 & 0.72 & 0.72 & 0.46 \\
\end{longtable}\endgroup

Reducing archive size does not restore the CMA result. However, neither
protocol matches the number of measurements per genotype: CMA averages
6.38 epoch appearances per genotype, versus about 1.11 for full uniform.
Protocol A also changes search duration and inherited-state maturity;
Protocol B disrupts temporal structure and leaves a single surviving
genotype in about 21.5\% of epochs, which receives the singleton Borda
reward. These are limits on a causal archive-size interpretation.

The full uniform archive has a higher global correlation between proxy and
truth than the CMA archive, despite worse selected accuracy. Its much wider
quality range and lower mean archive quality help explain why global rank
correlation alone is insufficient to diagnose top-of-pool selection.
In all 16 full trajectories, the Borda-selected genotype also has the most
epoch appearances. This is an observed property, not a general identity
of Borda selection.

\subsection{Repeated measurements}
\begingroup\scriptsize\setlength{\tabcolsep}{3pt}
\begin{longtable}{@{}lrrrrrr@{}}
\caption{Repeated-genotype analysis using raw stored loss. Correlations are oriented toward higher accuracy.}\label{tab:app-repetition}\\
\toprule
Subset & Cells & Single rho & Mean rho & Difference & t 95\% CI & Bootstrap 95\% CI \\
\midrule\endfirsthead
\multicolumn{7}{l}{\textit{Table \thetable{} continued.}}\\
\toprule
Subset & Cells & Single rho & Mean rho & Difference & t 95\% CI & Bootstrap 95\% CI \\
\midrule\endhead
\midrule\multicolumn{7}{r}{\textit{Continued on next page.}}\\\endfoot
\bottomrule\endlastfoot
All (primary) & 245 & 0.622 & 0.652 & 0.029 & [-0.001,0.059] & [-0.009,0.066] \\
Chance redraws & 124 & 0.718 & 0.772 & 0.055 & [0.018,0.091] & [0.004,0.102] \\
Incumbent-involved & 121 & 0.053 & 0.063 & 0.011 & [-0.175,0.197] & [-0.122,0.164] \\
\end{longtable}\endgroup

The repeated-genotype addendum contains 245 genotypes evaluated in at least
two epochs: 124 chance redraws and 121 with incumbent involvement. The
predeclared primary comparison, mean stored loss versus one randomly chosen
appearance, improves Spearman correlation by 0.029, but both the $t$ and
bootstrap intervals include zero. Chance redraws show a more consistent
secondary gain of about 0.055. Incumbent-involved observations are selected
by prior wins and use states influenced by those wins; they cannot be
treated as independent repeated measurements. These retrospective results
motivate, but do not replace, a prospective equal-evaluation experiment.

\section{CIFAR-100: Frozen Transfer and Calibration Depth}
\label{app:c100}
The final transfer result is the predeclared K6 arm of the five-level
CIFAR-100 study on $\mathcal S_8$. It is not the highest-scoring K selected
after examining the curve. Only the dataset, normalization, and classifier
output dimension differ from the frozen CIFAR-10 recipe.

\begingroup\scriptsize\setlength{\tabcolsep}{3pt}
\begin{longtable}{@{}lrrrrrrr@{}}
\caption{CIFAR-100 depth curve on $\mathcal S_8$. Test is report-only; K6 is the frozen transfer arm.}\label{tab:app-c100-agg}\\
\toprule
K & Validation & Test & Best & Regret & Unique & Spearman & Time (s) \\
\midrule\endfirsthead
\multicolumn{8}{l}{\textit{Table \thetable{} continued.}}\\
\toprule
K & Validation & Test & Best & Regret & Unique & Spearman & Time (s) \\
\midrule\endhead
\midrule\multicolumn{8}{r}{\textit{Continued on next page.}}\\\endfoot
\bottomrule\endlastfoot
0 & $69.452\pm0.773$ & $69.640\pm0.758$ & 71.485 & 2.033 & 153.5 & 0.323 & 204.2 \\
2 & $69.732\pm1.375$ & $69.657\pm1.152$ & 71.220 & 1.489 & 132.2 & 0.358 & 281.3 \\
4 & $69.390\pm1.539$ & $69.562\pm1.509$ & 71.467 & 2.077 & 138.6 & 0.371 & 339.8 \\
6 & $69.538\pm1.064$ & $69.615\pm1.139$ & 71.546 & 2.007 & 131.1 & 0.399 & 408.9 \\
8 & $69.755\pm1.437$ & $69.755\pm1.284$ & 71.978 & 2.223 & 143.2 & 0.370 & 472.0 \\
\end{longtable}\endgroup

\begingroup\scriptsize\setlength{\tabcolsep}{3pt}
\begin{longtable}{@{}lrrrrrr@{}}
\caption{CIFAR-100 selected validation by depth and K6 report-only test result.}\label{tab:app-c100-seed}\\
\toprule
Seed & K0 valid & K2 valid & K4 valid & K6 valid & K8 valid & K6 test \\
\midrule\endfirsthead
\multicolumn{7}{l}{\textit{Table \thetable{} continued.}}\\
\toprule
Seed & K0 valid & K2 valid & K4 valid & K6 valid & K8 valid & K6 test \\
\midrule\endhead
\midrule\multicolumn{7}{r}{\textit{Continued on next page.}}\\\endfoot
\bottomrule\endlastfoot
4611954 & 68.380 & 68.130 & 68.793 & 69.940 & 69.450 & 69.490 \\
4564271 & 70.880 & 70.880 & 69.947 & 69.947 & 69.860 & 70.700 \\
2621901 & 69.827 & 67.280 & 69.400 & 70.770 & 67.540 & 70.400 \\
1333636 & 69.553 & 69.947 & 71.293 & 69.860 & 69.480 & 70.027 \\
3046636 & 69.590 & 70.140 & 67.370 & 68.520 & 70.140 & 69.120 \\
3274845 & 69.080 & 71.440 & 71.440 & 69.960 & 69.340 & 70.200 \\
4096113 & 68.650 & 69.887 & 69.487 & 69.920 & 72.750 & 69.907 \\
7848893 & 69.660 & 70.150 & 67.390 & 67.390 & 69.480 & 67.080 \\
\end{longtable}\endgroup

\begingroup\scriptsize\setlength{\tabcolsep}{3pt}
\begin{longtable}{@{}lrrrrr@{}}
\caption{Paired CIFAR-100 validation contrasts; all intervals are exploratory.}\label{tab:app-c100-paired}\\
\toprule
Contrast & Mean & t 95\% CI & Bootstrap 95\% CI & W/T/L & Sign p \\
\midrule\endfirsthead
\multicolumn{6}{l}{\textit{Table \thetable{} continued.}}\\
\toprule
Contrast & Mean & t 95\% CI & Bootstrap 95\% CI & W/T/L & Sign p \\
\midrule\endhead
\midrule\multicolumn{6}{r}{\textit{Continued on next page.}}\\\endfoot
\bottomrule\endlastfoot
K2-K0 & 0.279 & [-0.891,1.450] & [-0.694,1.128] & 5/1/2 & 0.4531 \\
K4-K0 & -0.063 & [-1.496,1.371] & [-1.176,1.039] & 4/0/4 & 1.0000 \\
K6-K0 & 0.086 & [-1.050,1.221] & [-0.835,0.913] & 5/0/3 & 0.7266 \\
K8-K0 & 0.303 & [-1.242,1.847] & [-0.795,1.598] & 4/0/4 & 1.0000 \\
K6-K2 & -0.193 & [-1.879,1.493] & [-1.415,1.187] & 3/0/5 & 0.7266 \\
K6-K4 & 0.148 & [-0.787,1.084] & [-0.590,0.828] & 4/2/2 & 0.6875 \\
K6-K8 & -0.217 & [-1.823,1.389] & [-1.421,1.080] & 5/0/3 & 0.7266 \\
\end{longtable}\endgroup

At K6, selected validation is $69.538\pm1.064\%$ and report-only test
accuracy is $69.615\pm1.139\%$. The paired K6-minus-K0 validation change is
0.086 points with a $t$ interval of $[-1.050,1.221]$. Every displayed
K6-versus-other-depth interval includes zero. The highest mean validation
in the tested grid is at its upper boundary, K8, so these data do not show
an interior optimum. They also do not establish a calibration benefit on
CIFAR-100 comparable in precision to the CIFAR-10 replication.

Across the 40 searches, the source analysis records six fallback epochs at
K2 and one at K6; the other K arms have none. The K6 event belongs to seed
4096113. The earlier independent CIFAR-100 study used a different eight-seed
cohort and is reported only as historical context in
Appendix~\ref{app:development}.

\section{ImageNet16-120 and Matched Dataset Diagnostics}
\label{app:imagenet}
\begingroup\scriptsize\setlength{\tabcolsep}{3pt}
\begin{longtable}{@{}lrrrrrrr@{}}
\caption{ImageNet16-120 seed-level outcomes. Test is report-only; percentile is full-space midrank.}\label{tab:app-image-seed}\\
\toprule
Seed & Validation & Test & Percentile & Best & Regret & Unique & Time (s) \\
\midrule\endfirsthead
\multicolumn{8}{l}{\textit{Table \thetable{} continued.}}\\
\toprule
Seed & Validation & Test & Percentile & Best & Regret & Unique & Time (s) \\
\midrule\endhead
\midrule\multicolumn{8}{r}{\textit{Continued on next page.}}\\\endfoot
\bottomrule\endlastfoot
4611954 & 44.200 & 45.133 & 96.483 & 45.567 & 1.367 & 143 & 449.7 \\
4564271 & 44.683 & 44.533 & 98.122 & 46.183 & 1.500 & 172 & 453.2 \\
2621901 & 44.433 & 44.983 & 97.430 & 45.567 & 1.133 & 173 & 452.4 \\
1333636 & 41.778 & 42.267 & 84.835 & 45.633 & 3.856 & 149 & 453.3 \\
3046636 & 39.667 & 40.333 & 71.939 & 45.633 & 5.967 & 107 & 452.4 \\
3274845 & 44.678 & 45.022 & 98.090 & 46.033 & 1.356 & 183 & 453.2 \\
4096113 & 43.300 & 44.300 & 92.454 & 46.033 & 2.733 & 80 & 459.1 \\
7848893 & 42.700 & 43.189 & 89.469 & 46.089 & 3.389 & 137 & 451.2 \\
\end{longtable}\endgroup

\begingroup\scriptsize\setlength{\tabcolsep}{3pt}
\begin{longtable}{@{}lrrr@{}}
\caption{Matched eight-seed transfer diagnostics. Raw accuracy columns describe their own datasets only.}\label{tab:app-transfer}\\
\toprule
Metric & CIFAR-10 & CIFAR-100 & ImageNet16-120 \\
\midrule\endfirsthead
\multicolumn{4}{l}{\textit{Table \thetable{} continued.}}\\
\toprule
Metric & CIFAR-10 & CIFAR-100 & ImageNet16-120 \\
\midrule\endhead
\midrule\multicolumn{4}{r}{\textit{Continued on next page.}}\\\endfoot
\bottomrule\endlastfoot
Validation & $90.041\pm0.482$ & $69.538\pm1.064$ & $43.180\pm1.759$ \\
Test (report-only) & $93.388\pm0.367$ & $69.615\pm1.139$ & $43.720\pm1.697$ \\
Selected percentile & $93.52\pm3.44$ & $90.59\pm9.49$ & $91.10\pm9.09$ \\
Best-visited percentile & $97.24\pm1.90$ & $99.29\pm0.57$ & $99.73\pm0.14$ \\
Normalized oracle regret & $0.1977\pm0.0608$ & $0.3244\pm0.0873$ & $0.2749\pm0.1361$ \\
Spearman & $0.525\pm0.094$ & $0.399\pm0.176$ & $0.516\pm0.103$ \\
Kendall & $0.374\pm0.070$ & $0.278\pm0.126$ & $0.365\pm0.078$ \\
Unique genotypes & $132.9\pm45.4$ & $131.1\pm42.7$ & $143.0\pm35.2$ \\
Time (s) & $410.2\pm2.6$ & $408.9\pm9.3$ & $453.1\pm2.7$ \\
\end{longtable}\endgroup

\begingroup\scriptsize\setlength{\tabcolsep}{3pt}
\begin{longtable}{@{}lrrr@{}}
\caption{Normalized exploration/selection decomposition; entries are seed means unless noted.}\label{tab:app-transfer-gaps}\\
\toprule
Quantity & CIFAR-10 & CIFAR-100 & ImageNet16-120 \\
\midrule\endfirsthead
\multicolumn{4}{l}{\textit{Table \thetable{} continued.}}\\
\toprule
Quantity & CIFAR-10 & CIFAR-100 & ImageNet16-120 \\
\midrule\endhead
\midrule\multicolumn{4}{r}{\textit{Continued on next page.}}\\\endfoot
\bottomrule\endlastfoot
Exploration gap & 0.1191 & 0.1597 & 0.0689 \\
Selection gap & 0.0786 & 0.1646 & 0.2060 \\
Selection share of total gap & 0.398 & 0.508 & 0.749 \\
Seeds with selection gap larger & 2/8 & 4/8 & 7/8 \\
Best-visited Borda rank & 52.8 & 39.9 & 47.1 \\
Median selected truth rank in pool & 19.5 & 31.0 & 23.0 \\
\end{longtable}\endgroup

\begingroup\scriptsize\setlength{\tabcolsep}{3pt}
\begin{longtable}{@{}lrrrrrrr@{}}
\caption{Per-seed transfer diagnostics on the matched eight-seed cohort.}\label{tab:app-transfer-seed-diag}\\
\toprule
Dataset & Seed & Percentile & Regret & Unique & Spearman & Kendall & Best rank \\
\midrule\endfirsthead
\multicolumn{8}{l}{\textit{Table \thetable{} continued.}}\\
\toprule
Dataset & Seed & Percentile & Regret & Unique & Spearman & Kendall & Best rank \\
\midrule\endhead
\midrule\multicolumn{8}{r}{\textit{Continued on next page.}}\\\endfoot
\bottomrule\endlastfoot
CIFAR-10 & 4611954 & 95.17 & 1.401 & 161 & 0.424 & 0.293 & 5 \\
CIFAR-10 & 4564271 & 93.20 & 0.812 & 87 & 0.559 & 0.392 & 15 \\
CIFAR-10 & 2621901 & 89.58 & 0.482 & 104 & 0.636 & 0.464 & 57 \\
CIFAR-10 & 1333636 & 92.94 & 0.506 & 133 & 0.587 & 0.420 & 24 \\
CIFAR-10 & 3046636 & 94.86 & 0.208 & 97 & 0.504 & 0.366 & 16 \\
CIFAR-10 & 3274845 & 94.28 & 0.442 & 131 & 0.623 & 0.442 & 57 \\
CIFAR-10 & 4096113 & 99.59 & 0.491 & 229 & 0.491 & 0.349 & 139 \\
CIFAR-10 & 7848893 & 88.52 & 0.639 & 121 & 0.373 & 0.263 & 109 \\
CIFAR-100 & 4611954 & 94.84 & 0.880 & 124 & 0.399 & 0.280 & 57 \\
CIFAR-100 & 4564271 & 94.88 & 1.803 & 104 & 0.511 & 0.358 & 38 \\
CIFAR-100 & 2621901 & 98.49 & 1.977 & 213 & 0.290 & 0.202 & 102 \\
CIFAR-100 & 1333636 & 94.43 & 1.580 & 110 & 0.583 & 0.408 & 2 \\
CIFAR-100 & 3046636 & 82.00 & 2.170 & 182 & 0.337 & 0.228 & 35 \\
CIFAR-100 & 3274845 & 94.99 & 1.480 & 95 & 0.519 & 0.371 & 2 \\
CIFAR-100 & 4096113 & 94.70 & 2.380 & 117 & 0.510 & 0.351 & 27 \\
CIFAR-100 & 7848893 & 70.43 & 3.790 & 104 & 0.041 & 0.024 & 56 \\
ImageNet16-120 & 4611954 & 96.48 & 1.367 & 143 & 0.624 & 0.452 & 79 \\
ImageNet16-120 & 4564271 & 98.12 & 1.500 & 172 & 0.575 & 0.410 & 25 \\
ImageNet16-120 & 2621901 & 97.43 & 1.133 & 173 & 0.573 & 0.411 & 36 \\
ImageNet16-120 & 1333636 & 84.84 & 3.856 & 149 & 0.553 & 0.392 & 21 \\
ImageNet16-120 & 3046636 & 71.94 & 5.967 & 107 & 0.539 & 0.378 & 16 \\
ImageNet16-120 & 3274845 & 98.09 & 1.356 & 183 & 0.307 & 0.208 & 125 \\
ImageNet16-120 & 4096113 & 92.45 & 2.733 & 80 & 0.418 & 0.293 & 16 \\
ImageNet16-120 & 7848893 & 89.47 & 3.389 & 137 & 0.537 & 0.371 & 59 \\
\end{longtable}\endgroup
{\footnotesize Best rank is the Borda rank of the best-visited genotype, not its benchmark rank.}\par\medskip

All eight ImageNet16-120 searches completed, with no non-finite candidate
records and no fallback epochs. Selected validation is
$43.180\pm1.759\%$ and report-only test accuracy is
$43.720\pm1.697\%$. The mean selected full-space percentile is 91.10,
but two seeds return cells below the 85th percentile. Every run visits a
cell above the 99th percentile. The mean best-visited validation is 45.842,
and the mean selection regret is 2.662 points.

The normalized gap decomposition attributes roughly 75\% of the aggregate
ImageNet16-120 regret to selection within the visited pool. This fraction
is the ratio of the mean normalized selection gap to the mean normalized
total gap, not the mean of per-seed fractions. Seven of eight seeds have
a larger selection gap than exploration gap. Moderate overall Borda--truth
correlation therefore coexists with missed high-quality architectures.

Cross-dataset comparisons use percentiles, normalized regret, pool coverage,
and ranking correlations, not raw accuracy differences. The shared seed list
does not make the training problems identical. ImageNet16-120 costs more
overall because W25 trains on 151,700 examples, despite its cheaper
$16\times16$ search forwards. Its mean warm-up time is about 262.5 seconds
and its search time about 190.5 seconds, for 453.1 seconds of component time.

\section{Returned NAS-Bench-201 Architectures}
\label{app:nb-genotypes}
For compactness, each row lists operations in the fixed edge order
$(0,1)$, $(0,2)$, $(1,2)$, $(0,3)$, $(1,3)$, $(2,3)$.
The codes are N = \texttt{none}, S = \texttt{skip\_connect},
C1 = \texttt{nor\_conv\_1x1}, C3 = \texttt{nor\_conv\_3x3}, and
A3 = \texttt{avg\_pool\_3x3}. This lossless representation reconstructs
the canonical NB201 cell string. The tables include both primary paired
arms and the final transfer selections; all other ablation genotypes are
retained in the accompanying source tables.
\begingroup\scriptsize\setlength{\tabcolsep}{3pt}
\begin{longtable}{@{}lrrrrrrr@{}}
\caption{Returned NB201 genotypes. C10 I/P denotes initial/prospective cohort.}\label{tab:app-nb-genotypes}\\
\toprule
Cohort/arm & Seed & 0-1 & 0-2 & 1-2 & 0-3 & 1-3 & 2-3 \\
\midrule\endfirsthead
\multicolumn{8}{l}{\textit{Table \thetable{} continued.}}\\
\toprule
Cohort/arm & Seed & 0-1 & 0-2 & 1-2 & 0-3 & 1-3 & 2-3 \\
\midrule\endhead
\midrule\multicolumn{8}{r}{\textit{Continued on next page.}}\\\endfoot
\bottomrule\endlastfoot
C10 I K0 & 1333636 & C3 & A3 & C1 & C3 & C3 & S \\
C10 I K0 & 2621901 & C1 & C3 & C1 & C3 & N & N \\
C10 I K0 & 3011368 & N & C3 & C1 & C3 & C3 & S \\
C10 I K0 & 3046636 & C1 & C3 & S & C3 & C3 & S \\
C10 I K0 & 3274845 & C3 & S & N & C3 & A3 & C3 \\
C10 I K0 & 3458191 & S & A3 & C1 & C1 & C3 & C3 \\
C10 I K0 & 3750948 & C1 & C3 & C1 & C3 & C1 & C3 \\
C10 I K0 & 4096113 & C3 & C3 & A3 & S & C1 & S \\
C10 I K0 & 4564271 & C1 & C3 & C1 & C3 & C1 & C3 \\
C10 I K0 & 4611954 & C1 & C1 & C1 & C3 & C3 & N \\
C10 I K0 & 5197346 & C1 & C1 & C1 & C3 & C3 & C1 \\
C10 I K0 & 7848893 & C3 & C1 & S & C3 & A3 & S \\
C10 I K0 & 7990734 & C3 & S & S & C3 & S & C1 \\
C10 I K0 & 8224849 & S & C3 & C1 & A3 & C3 & C1 \\
C10 I K0 & 8629488 & N & N & A3 & C3 & S & S \\
C10 I K0 & 8682075 & C1 & C3 & C1 & C3 & S & C3 \\
C10 I K6 & 1333636 & C3 & C1 & C1 & C3 & C3 & C1 \\
C10 I K6 & 2621901 & C1 & C3 & C1 & C3 & C3 & C1 \\
C10 I K6 & 3011368 & S & C3 & C1 & C3 & C1 & C3 \\
C10 I K6 & 3046636 & C1 & C3 & C1 & C3 & C1 & C3 \\
C10 I K6 & 3274845 & C3 & C1 & C3 & C3 & C1 & C1 \\
C10 I K6 & 3458191 & S & C3 & C1 & C1 & C3 & C1 \\
C10 I K6 & 3750948 & C1 & S & C1 & C3 & C1 & C3 \\
C10 I K6 & 4096113 & C3 & C3 & C3 & S & S & C3 \\
C10 I K6 & 4564271 & C3 & C1 & C1 & C3 & C1 & C3 \\
C10 I K6 & 4611954 & C3 & C1 & C3 & C3 & S & C3 \\
C10 I K6 & 5197346 & S & C1 & C1 & C3 & C3 & C1 \\
C10 I K6 & 7848893 & C3 & C1 & C1 & C3 & A3 & C1 \\
C10 I K6 & 7990734 & C3 & C1 & C1 & C3 & N & C3 \\
C10 I K6 & 8224849 & S & C3 & C1 & C3 & C3 & N \\
C10 I K6 & 8629488 & C3 & S & C3 & C1 & C3 & C1 \\
C10 I K6 & 8682075 & C1 & C1 & C1 & C3 & C3 & C3 \\
C10 P K0 & 3297086 & S & C1 & S & C3 & S & C3 \\
C10 P K0 & 4007814 & C1 & C1 & C1 & C3 & S & C3 \\
C10 P K0 & 4450079 & S & C3 & A3 & C3 & S & C3 \\
C10 P K0 & 5392407 & C1 & S & S & C3 & C3 & C3 \\
C10 P K0 & 5805960 & C1 & C3 & C1 & C3 & A3 & S \\
C10 P K0 & 6963249 & C3 & C3 & A3 & C1 & C1 & S \\
C10 P K0 & 8147065 & N & C1 & A3 & C3 & C3 & C1 \\
C10 P K0 & 8167106 & C1 & S & C1 & C3 & N & C1 \\
C10 P K6 & 3297086 & S & C3 & C1 & C3 & S & C1 \\
C10 P K6 & 4007814 & C1 & C3 & C3 & C3 & C1 & C1 \\
C10 P K6 & 4450079 & S & C3 & C3 & C3 & S & C3 \\
C10 P K6 & 5392407 & C1 & S & C1 & C3 & S & C3 \\
C10 P K6 & 5805960 & C1 & C3 & C1 & C3 & C1 & C3 \\
C10 P K6 & 6963249 & C3 & C3 & C1 & C1 & C1 & C1 \\
C10 P K6 & 8147065 & C3 & S & N & S & C3 & C3 \\
C10 P K6 & 8167106 & C1 & C1 & C1 & C3 & C3 & C1 \\
C100 K6 & 1333636 & C3 & C1 & S & C3 & C1 & C1 \\
C100 K6 & 2621901 & C3 & S & S & C3 & C3 & S \\
C100 K6 & 3046636 & C1 & C3 & C3 & C3 & C1 & N \\
C100 K6 & 3274845 & C3 & C1 & C1 & C3 & C1 & C1 \\
C100 K6 & 4096113 & C3 & C3 & C1 & C3 & S & C1 \\
C100 K6 & 4564271 & C3 & C1 & N & C3 & C1 & C3 \\
C100 K6 & 4611954 & C1 & S & C1 & C3 & C3 & S \\
C100 K6 & 7848893 & S & C1 & S & C3 & C3 & S \\
IN16 K6 & 1333636 & S & C1 & S & C3 & C3 & C1 \\
IN16 K6 & 2621901 & C1 & S & C1 & C3 & C3 & C1 \\
IN16 K6 & 3046636 & S & C1 & C1 & C3 & A3 & C3 \\
IN16 K6 & 3274845 & C3 & C1 & N & C3 & C3 & S \\
IN16 K6 & 4096113 & C3 & C3 & C1 & C3 & C1 & C1 \\
IN16 K6 & 4564271 & C3 & C1 & C3 & C3 & C1 & S \\
IN16 K6 & 4611954 & C3 & S & C1 & C3 & C1 & S \\
IN16 K6 & 7848893 & C3 & C1 & S & C3 & C3 & C1 \\
\end{longtable}\endgroup

\section{DARTS Search-Space Adapter}
\label{app:darts-config}
\label{app:darts-adapter}
The DARTS adapter searches normal and reduction cells. Each cell has two
input nodes and four intermediate nodes; each intermediate node retains
exactly two incoming edges from distinct earlier nodes. Outputs of nodes
2--5 are concatenated. There are 14 candidate edges per cell and eight
latent operations per edge, giving $2\times14\times8=224$ latent logits.
The operations are \texttt{none}, max/average $3\times3$ pooling,
\texttt{skip\_connect}, separable $3\times3$/$5\times5$ convolutions, and
dilated $3\times3$/$5\times5$ convolutions.

After softmax and redistribution, each source edge is scored by its best
non-\texttt{none} operation. The decoder retains the two strongest distinct
sources at each node and chooses each retained edge's best non-null
operation. Source ties use source index and operation ties use primitive
order. \texttt{none} influences redistribution but is never an active
DARTS edge. The hard network contains no gradient-trained architecture
parameters and executes the decoded operations directly, without partial
channel sampling or edge-normalization weights.

\subsection{Topology-aware balance}
The exact NB201 per-edge coverage identity does not extend literally to
DARTS: a DARTS path activates only two sources per node, while other
candidate edges are absent. The implemented adapter therefore uses
\emph{approximate conditional operation balance} under the canonical
topology constraints. Its coverage state records counts for each cell,
candidate edge, and one of seven eligible non-null operations.

For a calibration schedule of K steps, all K focal-genotype occurrences
are reserved in the count tensor first. Each complement selects the two
least-used sources per node, then the least-used eligible operation on
each chosen edge, using seeded priorities to resolve ties. Counts are
updated after each complement. Every update contains the focal path plus
six valid complements, giving seven paths and 42 passes at K6. Warm-up
uses the same coverage mechanism with counts retained globally. The
coverage reports record actual edge and operation counts; this adapter
does not claim that all seven operations execute once on every one of
the 14 candidate edges in each update.

The remaining configuration---six updates, effective LR 0.056, the three
shared data panels, persistent pre-refresh state, CMA population and
scale, and final Borda rule---is unchanged. Forward-execution auditing
checks that active edges agree with the canonical decoded genotype.

\section{DARTS Retraining and Complete Per-Seed Results}
\label{app:darts-results}
\label{app:darts-tbd}
\begingroup\scriptsize\setlength{\tabcolsep}{3pt}
\begin{longtable}{@{}p{.43\linewidth}p{.49\linewidth}@{}}
\caption{Fixed DARTS retraining protocol, applied separately to each selected genotype.}\label{tab:app-darts-protocol}\\
\toprule
Setting & Value \\
\midrule\endfirsthead
\multicolumn{2}{l}{\textit{Table \thetable{} continued.}}\\
\toprule
Setting & Value \\
\midrule\endhead
\midrule\multicolumn{2}{r}{\textit{Continued on next page.}}\\\endfoot
\bottomrule\endlastfoot
Epochs / batch & 600 / 96 \\
Initial channels / cells & 36 / 20 \\
Training seed & 123 \\
Training / test examples & 50,000 / 10,000 \\
SGD LR / momentum / weight decay & 0.025 / 0.9 / 0.0003 \\
Scheduler / norm clipping & cosine to zero / 5 \\
Auxiliary head / loss weight & enabled / 0.4 \\
Cutout / drop-path maximum & 16 / 0.2 \\
Reported endpoint & best test accuracy over 600 epochs \\
Test evaluation & every epoch \\
Retraining cost & excluded from search time \\
\end{longtable}\endgroup

The selected cell is retrained from scratch on all 50,000 CIFAR training
images with a 20-cell, 36-channel network. The auxiliary classifier
contributes loss weight 0.4. Cutout has length 16; drop-path probability
increases as $0.2e/600$ for zero-based epoch $e$. SGD uses momentum 0.9,
weight decay $3\times10^{-4}$, initial LR 0.025 with cosine decay, and
gradient clipping at 5. The official 10,000-image test set is evaluated after each epoch.
For each architecture, we report the maximum test accuracy observed over
the 600-epoch trajectory and summarize these maxima across the three
architecture seeds. Architecture selection is completed before retraining.

\begingroup\scriptsize\setlength{\tabcolsep}{3pt}
\begin{longtable}{@{}lrrrrrrr@{}}
\caption{All six DARTS outcomes. Best accuracy over 600 epochs is the primary endpoint; final-epoch accuracy and error are supplementary.}\label{tab:app-darts-results}\\
\toprule
Dataset & Search seed & Final acc. & Final err. & Best acc. & Best epoch & Search min & Train h \\
\midrule\endfirsthead
\multicolumn{8}{l}{\textit{Table \thetable{} continued.}}\\
\toprule
Dataset & Search seed & Final acc. & Final err. & Best acc. & Best epoch & Search min & Train h \\
\midrule\endhead
\midrule\multicolumn{8}{r}{\textit{Continued on next page.}}\\\endfoot
\bottomrule\endlastfoot
CIFAR-10 & 184271 & 96.92 & 3.08 & 96.96 & 594 & 43.63 & 8.04 \\
CIFAR-10 & 392113 & 96.93 & 3.07 & 97.08 & 569 & 43.51 & 7.36 \\
CIFAR-10 & 618407 & 96.66 & 3.34 & 96.81 & 580 & 43.10 & 6.48 \\
CIFAR-100 & 184271 & 82.93 & 17.07 & 83.22 & 574 & 43.53 & 8.02 \\
CIFAR-100 & 392113 & 81.39 & 18.61 & 81.63 & 595 & 43.36 & 7.11 \\
CIFAR-100 & 618407 & 82.21 & 17.79 & 82.45 & 593 & 43.49 & 7.26 \\
\end{longtable}\endgroup
\begingroup\scriptsize\setlength{\tabcolsep}{3pt}
\begin{longtable}{@{}lrrrrr@{}}
\caption{DARTS results over three architecture seeds, each retrained with seed 123. Best accuracy over 600 epochs is the primary endpoint; epoch-600 results are supplementary.}\label{tab:app-darts-agg}\\
\toprule
Dataset & Epoch 600 acc. & Epoch 600 err. & Best acc. & Active min & GPU-days \\
\midrule\endfirsthead
\multicolumn{6}{l}{\textit{Table \thetable{} continued.}}\\
\toprule
Dataset & Final accuracy & Final error & Best diagnostic & Active min & GPU-days \\
\midrule\endhead
\midrule\multicolumn{6}{r}{\textit{Continued on next page.}}\\\endfoot
\bottomrule\endlastfoot
CIFAR-10 & $96.84\pm0.15$ & $3.16\pm0.15$ & $96.95\pm0.14$ & $43.41\pm0.28$ & 0.03015 \\
CIFAR-100 & $82.18\pm0.77$ & $17.82\pm0.77$ & $82.43\pm0.80$ & $43.46\pm0.09$ & 0.03018 \\
\end{longtable}\endgroup

\begingroup\scriptsize\setlength{\tabcolsep}{3pt}
\begin{longtable}{@{}lrrr@{}}
\caption{Reconstructed final-model parameter counts in millions, with counting conventions separated.}\label{tab:app-darts-model-size}\\
\toprule
Dataset & Search seed & Excluding auxiliary & Including auxiliary \\
\midrule\endfirsthead
\multicolumn{4}{l}{\textit{Table \thetable{} continued.}}\\
\toprule
Dataset & Search seed & Excluding auxiliary & Including auxiliary \\
\midrule\endhead
\midrule\multicolumn{4}{r}{\textit{Continued on next page.}}\\\endfoot
\bottomrule\endlastfoot
CIFAR-10 & 184271 & 3.5692 & 4.0457 \\
CIFAR-10 & 392113 & 3.2267 & 3.7031 \\
CIFAR-10 & 618407 & 2.9638 & 3.4402 \\
CIFAR-100 & 184271 & 3.6212 & 4.1668 \\
CIFAR-100 & 392113 & 3.2198 & 3.7655 \\
CIFAR-100 & 618407 & 3.3520 & 3.8977 \\
\end{longtable}\endgroup

Parameter counts are reconstructed from each stored genotype using the
released 36-channel, 20-cell retraining model. We give both the usual
count excluding the auxiliary head and the full training-model count,
so the counting convention is explicit.
Best-over-training and final-epoch results are reported separately because
they measure different endpoints. The main paper reports the former.
Each architecture is retrained once using seed 123; the reported standard
deviation therefore describes variation across searched architectures
under this fixed retraining seed, rather than retraining-seed variability.

\subsection{Search dynamics and memory}
\begingroup\scriptsize\setlength{\tabcolsep}{3pt}
\begin{longtable}{@{}lrrrrrrrr@{}}
\caption{Recorded DARTS search dynamics; peak memory is allocated GiB.}\label{tab:app-darts-dynamics}\\
\toprule
Dataset & Seed & Unique & Inc. wins & Transitions & Borda & Appearances & Fallback & Peak GiB \\
\midrule\endfirsthead
\multicolumn{9}{l}{\textit{Table \thetable{} continued.}}\\
\toprule
Dataset & Seed & Unique & Inc. wins & Transitions & Borda & Appearances & Fallback & Peak GiB \\
\midrule\endhead
\midrule\multicolumn{9}{r}{\textit{Continued on next page.}}\\\endfoot
\bottomrule\endlastfoot
CIFAR-10 & 184271 & 578 & 59 & 1 & 60.000 & 60 & 0 & 4.29 \\
CIFAR-10 & 392113 & 667 & 37 & 23 & 29.917 & 30 & 0 & 4.28 \\
CIFAR-10 & 618407 & 566 & 26 & 34 & 14.333 & 17 & 0 & 4.30 \\
CIFAR-100 & 184271 & 510 & 59 & 1 & 60.000 & 60 & 0 & 4.16 \\
CIFAR-100 & 392113 & 514 & 59 & 1 & 60.000 & 60 & 0 & 4.20 \\
CIFAR-100 & 618407 & 574 & 37 & 23 & 20.583 & 22 & 0 & 4.22 \\
\end{longtable}\endgroup

All six selected runs contain 60 search epochs and 780 candidate events;
no proposal fallback was used. Incumbent persistence varies substantially
across seeds. The same genotype is selected on CIFAR-10 and CIFAR-100 for
architecture seed 184271, although the dataset-specific searches and
retrainings are separate. This repeated selection is visible in the
canonical cells below and should not be counted as six distinct cell pairs.

\subsection{Normal and reduction cells}
For each intermediate node, a pair $(o,s)$ denotes operation $o$ from
source node $s$. Codes are S3/S5 for separable convolutions, D3/D5 for
dilated convolutions, M3/A3 for max/average pooling, and I for identity.
All cells concatenate nodes $[2,3,4,5]$.
\begingroup\scriptsize\setlength{\tabcolsep}{3pt}
\begin{longtable}{@{}lrrll@{}}
\caption{Complete normal and reduction cells for every reported DARTS search.}\label{tab:app-darts-genotypes}\\
\toprule
Dataset & Seed & Node & Normal incoming pairs & Reduction incoming pairs \\
\midrule\endfirsthead
\multicolumn{5}{l}{\textit{Table \thetable{} continued.}}\\
\toprule
Dataset & Seed & Node & Normal incoming pairs & Reduction incoming pairs \\
\midrule\endhead
\midrule\multicolumn{5}{r}{\textit{Continued on next page.}}\\\endfoot
\bottomrule\endlastfoot
CIFAR-10 & 184271 & 2 & (S5,0), (M3,1) & (A3,0), (A3,1) \\
CIFAR-10 & 184271 & 3 & (S5,0), (S3,1) & (S3,0), (S3,1) \\
CIFAR-10 & 184271 & 4 & (D5,0), (S3,1) & (S3,0), (A3,2) \\
CIFAR-10 & 184271 & 5 & (M3,3), (D5,4) & (S3,0), (D3,1) \\
CIFAR-10 & 392113 & 2 & (D5,0), (M3,1) & (M3,0), (M3,1) \\
CIFAR-10 & 392113 & 3 & (A3,0), (D5,2) & (S5,0), (S3,2) \\
CIFAR-10 & 392113 & 4 & (S5,0), (M3,2) & (D5,1), (S3,3) \\
CIFAR-10 & 392113 & 5 & (S5,1), (S3,2) & (A3,0), (S5,2) \\
CIFAR-10 & 618407 & 2 & (S3,0), (D3,1) & (A3,0), (D5,1) \\
CIFAR-10 & 618407 & 3 & (D5,1), (D3,2) & (S5,0), (D5,1) \\
CIFAR-10 & 618407 & 4 & (A3,1), (A3,3) & (S5,0), (D5,1) \\
CIFAR-10 & 618407 & 5 & (D3,1), (D5,3) & (A3,2), (S5,3) \\
CIFAR-100 & 184271 & 2 & (S5,0), (M3,1) & (A3,0), (A3,1) \\
CIFAR-100 & 184271 & 3 & (S5,0), (S3,1) & (S3,0), (S3,1) \\
CIFAR-100 & 184271 & 4 & (D5,0), (S3,1) & (S3,0), (A3,2) \\
CIFAR-100 & 184271 & 5 & (M3,3), (D5,4) & (S3,0), (D3,1) \\
CIFAR-100 & 392113 & 2 & (D5,0), (M3,1) & (D3,0), (D3,1) \\
CIFAR-100 & 392113 & 3 & (S3,1), (D5,2) & (S5,0), (S3,2) \\
CIFAR-100 & 392113 & 4 & (S5,0), (M3,2) & (D3,1), (M3,3) \\
CIFAR-100 & 392113 & 5 & (M3,0), (S3,4) & (S3,0), (M3,3) \\
CIFAR-100 & 618407 & 2 & (S3,0), (D3,1) & (M3,0), (D5,1) \\
CIFAR-100 & 618407 & 3 & (S3,1), (S3,2) & (S5,0), (D3,1) \\
CIFAR-100 & 618407 & 4 & (A3,1), (A3,3) & (S5,0), (D5,1) \\
CIFAR-100 & 618407 & 5 & (A3,0), (S3,1) & (D3,0), (S5,3) \\
\end{longtable}\endgroup

\section{Runtime Boundaries, Hardware, and Reproducibility}
\label{app:reproducibility}
\begingroup\scriptsize\setlength{\tabcolsep}{3pt}
\begin{longtable}{@{}lrrrrrr@{}}
\caption{Runtime decomposition in seconds. NB201 K0 timing-only exclusions do not exclude accuracy observations.}\label{tab:app-runtime}\\
\toprule
Cohort & Arm & Timing n & Warm-up & Search & Selection & Complete \\
\midrule\endfirsthead
\multicolumn{7}{l}{\textit{Table \thetable{} continued.}}\\
\toprule
Cohort & Arm & Timing n & Warm-up & Search & Selection & Complete \\
\midrule\endhead
\midrule\multicolumn{7}{r}{\textit{Continued on next page.}}\\\endfoot
\bottomrule\endlastfoot
Initial 16 & K6 & 16 & 99.7 & 310.7 & 0.0200 & 410.4 \\
Initial 16 & K0 & 14 & 99.7 & 106.0 & 0.0187 & 205.8 \\
Prospective 8 & K6 & 8 & 99.5 & 307.0 & 0.0196 & 406.4 \\
Prospective 8 & K0 & 8 & 99.5 & 105.2 & 0.0181 & 204.7 \\
Pooled 24 & K6 & 24 & 99.6 & 309.5 & 0.0198 & 409.1 \\
Pooled 24 & K0 & 22 & 99.6 & 105.7 & 0.0185 & 205.4 \\
CIFAR-100 & K6 & 8 & 99.4 & 309.4 & 0.0199 & 408.9 \\
ImageNet16-120 & K6 & 8 & 262.5 & 190.5 & 0.0204 & 453.1 \\
DARTS CIFAR-10 & K6 & 3 & 791.2 & 1813.5 & 0.0225 & 2604.7 \\
DARTS CIFAR-100 & K6 & 3 & 790.9 & 1816.5 & 0.0224 & 2607.5 \\
\end{longtable}\endgroup

\begingroup\scriptsize\setlength{\tabcolsep}{3pt}
\begin{longtable}{@{}lrrr@{}}
\caption{Instrumented search components, mean seconds over matched transfer seeds.}\label{tab:app-transfer-runtime}\\
\toprule
Component & CIFAR-10 & CIFAR-100 & ImageNet16-120 \\
\midrule\endfirsthead
\multicolumn{4}{l}{\textit{Table \thetable{} continued.}}\\
\toprule
Component & CIFAR-10 & CIFAR-100 & ImageNet16-120 \\
\midrule\endhead
\midrule\multicolumn{4}{r}{\textit{Continued on next page.}}\\\endfoot
\bottomrule\endlastfoot
Search initialization & 0.63 & 0.79 & 1.59 \\
Panel loading & 33.98 & 32.23 & 29.36 \\
Panel fingerprints & 2.34 & 2.36 & 0.84 \\
Candidate calibration & 195.66 & 195.72 & 94.10 \\
Focal BN refresh & 28.18 & 26.46 & 15.68 \\
Validation scoring & 9.52 & 9.07 & 9.13 \\
Proposal generation & 0.63 & 3.31 & 0.77 \\
\end{longtable}\endgroup
{\footnotesize These measured subcomponents do not exhaust process time; state cloning, logging, diagnostics, and other overhead remain.}\par\medskip

The NB201 analysis tables use warm-up training wall time plus search and
recorded selection time. Reused checkpoints are charged their recorded
warm-up cost rather than assigned zero cost. The runner's end-to-end
manifest time can additionally include process startup and dataset
initialization; for ImageNet16-120 its mean is 455.8 seconds versus
453.1 seconds for the component sum. These quantities use different timing boundaries and are reported separately. Search time excludes dataset download and the
benchmark's independent training-from-scratch evaluations.

DARTS \texttt{runtime.json} defines active warm-up plus search epochs,
including event and checkpoint writes, plus validated Borda selection.
It excludes setup, resume downtime, and retraining. Its approximately
43.4-minute mean active cost is a per-run quantity; queue elapsed time
and concurrent-job overlap are not substituted for it. GPU-days are
seconds divided by 86,400 for one GPU. The separate 600-epoch
retrainings take hours and are not part of the search-time comparison.

The recorded final runs use an NVIDIA GeForce RTX 4090 with 24 GB memory,
Python 3.10.20, PyTorch 2.5.1 with CUDA 12.1, NumPy 2.2.6, and
\texttt{cmaes} 0.13.0. The NB201 environment additionally records
torchvision 0.20.1 and driver 580.173.02. Determinism settings seed
Python, NumPy, PyTorch, and CUDA; disable cuDNN benchmarking; and set
\texttt{CUBLAS\_WORKSPACE\_CONFIG} to \texttt{:4096:8}. Fixed BLAS thread counts
matter for bitwise CMA eigendecomposition replay.

\subsection{Verification records and implementation equivalence}
The final search artifacts record configurations, schedule hashes,
candidate losses, decoded genotypes, pre-refresh inherited-state hashes,
proposal rejection statistics, and selected-genotype records. Completion
checks require the full candidate count and a replayable Borda selection.
Search processes install a benchmark-access guard; post-hoc truth joins
operate on completed outputs. DARTS additionally rejects official test
file access during search and audits hard-path topology.

The preserved E120 reproduction gate compares a 60-epoch prefix with a
historical run from the same W25 checkpoint: 60 schedule rows, 2,170 raw
proposal records, and 780 ordered candidate records agree, including
losses, winners, inherited hashes, and selection. The dense K/W analysis
rebuilds all 112 schedules, replays all selectors, and verifies the
shared-checkpoint and panel-prefix invariants. Its compatibility checks
bridge the recorded 81d09fd and ed71916 revisions without treating the
same reference runs as independent observations. The ImageNet16-120
analysis records a separate pre-truth seal and verifies all eight runs.

DARTS uses epoch transaction directories and checkpoints containing
weights, CMA state and RNG, incumbent vector, configuration, and RNG
streams. Warm-up checkpoints also store optimizer, scheduler, and
sampler state. Interrupted suffixes are archived and a validated
checkpoint is resumed; failed or superseded retraining attempts do not
enter the six final result rows. This recovery mechanism supports
execution continuity but does not make wall time spent in abandoned
attempts part of the active-runtime metric.

\section{Earlier Development Evidence Kept Separate}
\label{app:development}
These tables document available current-repository development evidence
without adding it to the principal sample size. No proxy-pilot analysis
is used. Historical folder names such as ``final analysis'' do not
override their actual seed lists, schedule generations, and code revisions.

\subsection{Earlier 16-pair NB201 cohort}
\begingroup\scriptsize\setlength{\tabcolsep}{3pt}
\begin{longtable}{@{}lrrrrrr@{}}
\caption{Earlier development pairs, excluded from the current 24-seed principal synthesis.}\label{tab:app-legacy}\\
\toprule
Block & Seed & K0 valid & K6 valid & Difference & K0 test & K6 test \\
\midrule\endfirsthead
\multicolumn{7}{l}{\textit{Table \thetable{} continued.}}\\
\toprule
Block & Seed & K0 valid & K6 valid & Difference & K0 test & K6 test \\
\midrule\endhead
\midrule\multicolumn{7}{r}{\textit{Continued on next page.}}\\\endfoot
\bottomrule\endlastfoot
original & 224307 & 89.314 & 91.072 & 1.758 & 92.415 & 93.833 \\
original & 283331 & 89.593 & 90.024 & 0.431 & 93.087 & 93.425 \\
original & 524955 & 89.654 & 90.063 & 0.409 & 92.990 & 93.137 \\
original & 544369 & 88.731 & 91.116 & 2.385 & 92.140 & 94.020 \\
original & 557081 & 89.856 & 91.072 & 1.216 & 93.415 & 93.833 \\
original & 636408 & 90.778 & 89.471 & -1.307 & 93.730 & 93.160 \\
original & 671413 & 90.466 & 89.446 & -1.020 & 93.210 & 93.020 \\
original & 738731 & 89.124 & 89.830 & 0.706 & 92.260 & 93.565 \\
manual\_extension & 1203987 & 88.794 & 90.010 & 1.216 & 91.940 & 92.815 \\
manual\_extension & 2317741 & 89.900 & 90.094 & 0.194 & 93.315 & 93.150 \\
manual\_extension & 3420919 & 88.768 & 89.591 & 0.823 & 91.940 & 92.680 \\
manual\_extension & 4576283 & 89.534 & 88.964 & -0.570 & 93.360 & 92.337 \\
manual\_extension & 5680447 & 88.662 & 89.210 & 0.548 & 92.065 & 92.685 \\
manual\_extension & 6793511 & 89.912 & 90.904 & 0.992 & 93.080 & 93.710 \\
manual\_extension & 7849207 & 89.238 & 89.774 & 0.536 & 92.685 & 93.170 \\
manual\_extension & 8956133 & 90.144 & 90.094 & -0.050 & 93.250 & 93.150 \\
\end{longtable}\endgroup

This historical cohort consists of an original eight-seed block and a
manual eight-seed extension, predating the principal v4 cohort. The
manual pairs regenerated warm-up files: serialized hashes differed,
but the independent verification found all 338 model-state tensors
byte-identical within every pair. The blocks also span different
recorded revisions. Their pooled validation gain is 0.517 points,
with $t$ interval $[0.008,1.026]$, bootstrap interval $[0.058,0.967]$,
and 12 wins/four losses (sign $p=0.0768$). This is developmental
support and is not combined with the newer 24-seed synthesis.

\subsection{Earlier CIFAR-100 transfer cohort}
\begingroup\scriptsize\setlength{\tabcolsep}{3pt}
\begin{longtable}{@{}lrrrrrr@{}}
\caption{Earlier CIFAR-100 transfer cohort, distinct from the final depth-study seeds.}\label{tab:app-legacy-c100}\\
\toprule
Seed & Validation & Test & Best & Regret & Unique & Time (s) \\
\midrule\endfirsthead
\multicolumn{7}{l}{\textit{Table \thetable{} continued.}}\\
\toprule
Seed & Validation & Test & Best & Regret & Unique & Time (s) \\
\midrule\endhead
\midrule\multicolumn{7}{r}{\textit{Continued on next page.}}\\\endfoot
\bottomrule\endlastfoot
738731 & 69.350 & 69.630 & 71.440 & 2.090 & 91 & 410.5 \\
283331 & 68.800 & 69.230 & 71.680 & 2.880 & 124 & 405.3 \\
224307 & 69.360 & 69.153 & 71.490 & 2.130 & 131 & 405.5 \\
544369 & 67.910 & 68.040 & 71.490 & 3.580 & 138 & 401.5 \\
671413 & 68.873 & 67.960 & 71.750 & 2.877 & 109 & 403.5 \\
636408 & 69.280 & 69.360 & 71.490 & 2.210 & 109 & 415.6 \\
557081 & 69.480 & 69.710 & 70.990 & 1.510 & 113 & 407.4 \\
524955 & 70.140 & 69.850 & 71.113 & 0.973 & 100 & 411.6 \\
\end{longtable}\endgroup

The historical eight-seed CIFAR-100 transfer used the original
development seed list, not $\mathcal S_8$. One initially failed attempt
at seed 636408 was preserved and excluded; its completed replacement
used the deterministic fallback once, at epoch 31, with 11 unique
proposals and one duplicate slot. The completed-run statistics above
are kept distinct from the final K6 depth-study transfer estimate.

\subsection{CMA-scale development and alternative-scale study}
\begingroup\scriptsize\setlength{\tabcolsep}{3pt}
\begin{longtable}{@{}lrrrrrrr@{}}
\caption{Complete four-seed CMA-scale development grid. Validation, not test, informed development.}\label{tab:app-sigma}\\
\toprule
Seed & Sigma & Validation & Test & Best & Regret & Unique & Fallback \\
\midrule\endfirsthead
\multicolumn{8}{l}{\textit{Table \thetable{} continued.}}\\
\toprule
Seed & Sigma & Validation & Test & Best & Regret & Unique & Fallback \\
\midrule\endhead
\midrule\multicolumn{8}{r}{\textit{Continued on next page.}}\\\endfoot
\bottomrule\endlastfoot
1203987 & 0.30 & 90.010 & 92.815 & 90.392 & 0.382 & 148 & 0 \\
1203987 & 0.35 & 89.956 & 93.020 & 91.106 & 1.150 & 148 & 0 \\
1203987 & 0.40 & 88.896 & 92.395 & 90.841 & 1.945 & 168 & 0 \\
1203987 & 0.45 & 90.305 & 93.080 & 90.656 & 0.351 & 146 & 0 \\
1203987 & 0.50 & 89.284 & 92.205 & 91.106 & 1.822 & 141 & 0 \\
1203987 & 0.55 & 89.516 & 92.890 & 91.188 & 1.672 & 139 & 0 \\
2317741 & 0.30 & 90.094 & 93.150 & 91.418 & 1.324 & 191 & 0 \\
2317741 & 0.35 & 89.293 & 92.417 & 90.485 & 1.192 & 138 & 0 \\
2317741 & 0.40 & 90.085 & 93.630 & 90.712 & 0.627 & 145 & 0 \\
2317741 & 0.45 & 89.900 & 93.315 & 91.324 & 1.424 & 147 & 0 \\
2317741 & 0.50 & 91.418 & 93.795 & 91.607 & 0.189 & 163 & 0 \\
2317741 & 0.55 & 90.053 & 93.210 & 91.533 & 1.480 & 136 & 0 \\
3420919 & 0.30 & 89.591 & 92.680 & 90.474 & 0.883 & 132 & 0 \\
3420919 & 0.35 & 90.013 & 92.780 & 91.119 & 1.105 & 126 & 0 \\
3420919 & 0.40 & 88.651 & 91.873 & 90.493 & 1.843 & 176 & 0 \\
3420919 & 0.45 & 90.474 & 93.545 & 90.474 & 0.000 & 144 & 0 \\
3420919 & 0.50 & 91.072 & 93.833 & 91.366 & 0.294 & 187 & 0 \\
3420919 & 0.55 & 89.116 & 92.003 & 91.132 & 2.016 & 201 & 0 \\
4576283 & 0.30 & 88.964 & 92.337 & 90.647 & 1.683 & 182 & 0 \\
4576283 & 0.35 & 90.094 & 93.150 & 90.485 & 0.391 & 166 & 0 \\
4576283 & 0.40 & 89.666 & 93.075 & 90.647 & 0.981 & 126 & 0 \\
4576283 & 0.45 & 91.106 & 93.760 & 91.106 & 0.000 & 167 & 0 \\
4576283 & 0.50 & 89.897 & 92.697 & 91.132 & 1.235 & 228 & 0 \\
4576283 & 0.55 & 90.956 & 93.820 & 91.607 & 0.651 & 219 & 0 \\
\end{longtable}\endgroup

The four-seed sigma grid favored 0.45 on its development validation
mean, but its paired improvement over 0.30 was imprecise. This led to
a separate v4 study at $\sigma=0.45$, rather than establishing a
replacement for the final 0.30 setting.

\begingroup\scriptsize\setlength{\tabcolsep}{3pt}
\begin{longtable}{@{}lrrrrrr@{}}
\caption{Separate 16-seed v4 study at initial CMA sigma 0.45; not pooled with sigma 0.30.}\label{tab:app-sigma045}\\
\toprule
Seed & K0 valid & K6 valid & Difference & K6 test & K6 best & K6 regret \\
\midrule\endfirsthead
\multicolumn{7}{l}{\textit{Table \thetable{} continued.}}\\
\toprule
Seed & K0 valid & K6 valid & Difference & K6 test & K6 best & K6 regret \\
\midrule\endhead
\midrule\multicolumn{7}{r}{\textit{Continued on next page.}}\\\endfoot
\bottomrule\endlastfoot
3170282 & 88.838 & 89.345 & 0.507 & 92.600 & 90.712 & 1.367 \\
3981128 & 88.426 & 89.196 & 0.770 & 92.920 & 91.237 & 2.041 \\
7054363 & 90.544 & 91.106 & 0.562 & 93.760 & 91.106 & 0.000 \\
4010385 & 90.902 & 89.108 & -1.794 & 92.020 & 90.364 & 1.256 \\
3428960 & 90.904 & 90.986 & 0.082 & 93.655 & 91.607 & 0.621 \\
6505985 & 89.132 & 89.336 & 0.204 & 92.235 & 91.106 & 1.770 \\
1564295 & 89.052 & 89.816 & 0.764 & 93.127 & 91.072 & 1.256 \\
2190322 & 87.335 & 90.393 & 3.059 & 93.303 & 91.154 & 0.761 \\
4442649 & 88.633 & 89.424 & 0.791 & 93.167 & 91.488 & 2.064 \\
1237366 & 88.624 & 89.052 & 0.428 & 92.635 & 91.237 & 2.185 \\
3359507 & 90.769 & 89.234 & -1.535 & 92.105 & 91.533 & 2.299 \\
3024771 & 89.924 & 90.743 & 0.819 & 93.613 & 91.106 & 0.363 \\
7306832 & 89.134 & 89.660 & 0.526 & 93.000 & 90.410 & 0.750 \\
7034536 & 88.484 & 89.735 & 1.251 & 93.113 & 91.250 & 1.515 \\
1036077 & 88.696 & 89.612 & 0.916 & 93.080 & 90.485 & 0.873 \\
8940720 & 89.589 & 89.510 & -0.079 & 92.920 & 90.540 & 1.030 \\
\end{longtable}\endgroup

At $\sigma=0.45$, the 16-seed mean K6-minus-K0 validation change is
0.454 points, with $t$ interval $[-0.123,1.031]$ and bootstrap interval
$[-0.062,0.968]$. Thirteen of 16 differences are positive (sign
$p=0.0213$), illustrating the difference between evidence about
direction and precision about the mean magnitude. These seeds,
configuration, and endpoints remain separate from the frozen
$\sigma=0.30$ evidence. The eight-seed secondary depth curve is also
reported for completeness.
\begingroup\scriptsize\setlength{\tabcolsep}{3pt}
\begin{longtable}{@{}lrrrrrr@{}}
\caption{Secondary eight-seed depth curve at sigma 0.45. Runtime is search-only, not complete-method time.}\label{tab:app-sigma045-depth}\\
\toprule
K & Validation & Best & Regret & Unique & Spearman & Search (s) \\
\midrule\endfirsthead
\multicolumn{7}{l}{\textit{Table \thetable{} continued.}}\\
\toprule
K & Validation & Best & Regret & Unique & Spearman & Search (s) \\
\midrule\endhead
\midrule\multicolumn{7}{r}{\textit{Continued on next page.}}\\\endfoot
\bottomrule\endlastfoot
0 & $89.392\pm1.284$ & $90.738\pm0.490$ & $1.346\pm0.923$ & $207.4\pm27.6$ & $0.372\pm0.125$ & $101.3\pm0.9$ \\
2 & $89.344\pm1.063$ & $90.736\pm0.618$ & $1.391\pm0.681$ & $188.6\pm28.1$ & $0.293\pm0.140$ & $173.1\pm2.7$ \\
4 & $89.828\pm0.806$ & $91.139\pm0.469$ & $1.311\pm0.626$ & $158.8\pm35.4$ & $0.467\pm0.073$ & $241.2\pm3.0$ \\
6 & $89.911\pm0.813$ & $91.045\pm0.368$ & $1.134\pm0.655$ & $156.6\pm31.2$ & $0.463\pm0.097$ & $306.4\pm6.7$ \\
8 & $89.919\pm0.975$ & $91.130\pm0.491$ & $1.211\pm0.587$ & $186.9\pm51.7$ & $0.395\pm0.118$ & $370.8\pm2.9$ \\
\end{longtable}\endgroup

\section{Evidence Provenance and Limits of Interpretation}
\label{app:provenance}
\begingroup\scriptsize\setlength{\tabcolsep}{3pt}
\begin{longtable}{@{}p{.23\linewidth}p{.70\linewidth}@{}}
\caption{Artifact families used to construct the appendix.}\label{tab:app-sources}\\
\toprule
Family & Authoritative artifact scope \\
\midrule\endfirsthead
\multicolumn{2}{l}{\textit{Table \thetable{} continued.}}\\
\toprule
Family & Authoritative artifact scope \\
\midrule\endhead
\midrule\multicolumn{2}{r}{\textit{Continued on next page.}}\\\endfoot
\bottomrule\endlastfoot
Principal CIFAR-10 & Final evidence suite: initial16, prospective8, matched controls, selector replay, runtime tables. \\
Dense K/W & Merged September 15 analysis: 112 compatible runs; K=0--8 and W=5,10,15,25,35,50. \\
Conditioning control & September 21 candidate-independent fair-calibration analysis, eight matched pairs. \\
CIFAR-100 & September 9 depth-study per-run rows and aggregate/paired tables. \\
ImageNet16-120 & Independent September 14 transfer analysis and matched dataset diagnostics. \\
Longer search & Independent E120 budget analysis: prefix and block tables. \\
Uniform archives & September 13 archive-size and repetition reanalysis of frozen logs. \\
DARTS & Final 4090 v4 September 11 run tree: six search outputs, seed-123 retraining results and 600-epoch logs. \\
Earlier development & Independent earlier final/transfer analyses and current-repository sigma studies; kept separate. \\
\end{longtable}\endgroup

Tables in this appendix were constructed from the completed-run and
analysis artifacts summarized above. Aggregate values were transcribed
from machine-readable summaries or recomputed from per-seed records.

The principal limitations are the small seed counts for ablations and
transfer, repeated reuse of $\mathcal S_8$, unadjusted exploratory
comparisons, possible dependence between evolving inherited weights
and repeated genotype evaluations, and the distinction between
benchmark averages and a fresh final training run. The candidate-
independent control has a skewed paired effect. Dense K/W curves do
not establish a unique optimum. Longer search yields diminishing
discovery and uncertain selected-accuracy gains. On ImageNet16-120,
high best-visited accuracy coexists with substantial selection regret.
DARTS balance is topology-constrained and approximate, and its
three-architecture estimates use only one final training seed.
These qualifications bound the claims supported by the available
experiments without changing the frozen method or excluding
unfavorable completed runs.